\documentclass[final,times,twocolumn]{elsarticle} %final
\usepackage{medima}
\usepackage{framed,multirow}

\usepackage{latexsym}

\usepackage{url}
\usepackage{xcolor}
\usepackage{cite}

\usepackage{amsmath,amssymb,amsfonts}
\usepackage{algorithmic}
\usepackage{graphicx}
\usepackage{epsfig}
\usepackage{epstopdf}
\usepackage{textcomp}

\usepackage{booktabs}
\usepackage{diagbox}
\usepackage{tabularx}
\usepackage{setspace}
\usepackage{color}
\usepackage[colorlinks,linkcolor=blue]{hyperref}
\usepackage{cleveref}
\usepackage{multirow}
\usepackage{tablefootnote}
\usepackage{threeparttable}
\usepackage{url}
\usepackage{caption}
\usepackage{bm}

\usepackage{amsmath}
\usepackage{amssymb}
\usepackage{pifont}
\usepackage{enumitem}

\usepackage{longtable}

\definecolor{newcolor}{rgb}{.8,.349,.1}

\journal{Medical Image Analysis}

\begin{document}

\verso{Yuhao Huang \textit{et~al.}}

\begin{frontmatter}

\title{Flip Learning: Weakly Supervised Erase to Segment Nodules in Breast Ultrasound}%

\author[1,2,3]{Yuhao \snm{Huang}}
% \ead{huangyuhao2019@email.szu.edu.cn}
\author[1,2,3]{Ao \snm{Chang}}
\author[4,7]{Haoran \snm{Dou}}
\author[1,2,3]{Xing \snm{Tao}}
\author[1,2,3]{Xinrui \snm{Zhou}}
\author[5]{Yan \snm{Cao}}
\author[1,2,3]{Ruobing \snm{Huang}}
\author[6,7,8,9,10]{Alejandro F. \snm{Frangi}}
\author[11]{Lingyun \snm{Bao}\corref{cor1}}
\ead{baolingyun2021@163.com}
\author[1,2,3]{Xin \snm{Yang}\corref{cor1}}
\ead{xinyang@szu.edu.cn}
\author[1,2,3,12]{Dong \snm{Ni}\corref{cor1}}
\ead{nidong@szu.edu.cn}

\cortext[cor1]{Corresponding authors.}

\address[1]{National-Regional Key Technology Engineering Laboratory for Medical Ultrasound, School of Biomedical Engineering, Shenzhen University Medical School, Shenzhen University, Shenzhen, China}
\address[2]{Medical UltraSound Image Computing (MUSIC) Lab, Shenzhen University, Shenzhen, China}
\address[3]{Marshall Laboratory of Biomedical Engineering, Shenzhen University, Shenzhen, China}
% \address[4]{Centre for Computational Imaging and Simulation Technologies in Biomedicine (CISTIB), University of Leeds, Leeds, UK}
\address[4]{School of Computing, University of Leeds, Leeds, UK}
\address[5]{Shenzhen RayShape Medical Technology Co., Ltd, Shenzhen, China}
\address[6]{Division of Informatics, Imaging and Data Science, School of Health Sciences, University of Manchester, Manchester, UK}
\address[7]{Department of Computer Science, School of Engineering, University of Manchester, Manchester, UK}
\address[8]{Medical Imaging Research Center (MIRC), Department of Electrical
Engineering, Department of Cardiovascular Sciences, KU Leuven, Belgium}
\address[9]{Alan Turing Institute, London, UK}
\address[10]{NIHR Manchester Biomedical Research Centre,
Manchester Academic Health Science Centre, Manchester, UK}
\address[11]{Department of Ultrasound, Affiliated Hangzhou First People's Hospital, School of Medicine, Westlake University}
\address[12]{School of Biomedical Engineering and Informatics, Nanjing Medical University, Nanjing, China}

\received{*****}
\finalform{*****}
\accepted{*****}
\availableonline{*****}
\communicated{S. Sarkar}

\begin{abstract}
Accurate segmentation of nodules in both 2D breast ultrasound (BUS) and 3D automated breast ultrasound (ABUS) is crucial for clinical diagnosis and treatment planning. 
Therefore, developing an automated system for nodule segmentation can enhance user independence and expedite clinical analysis. 
Unlike fully-supervised learning, weakly-supervised segmentation (WSS) can streamline the laborious and intricate annotation process.
However, current WSS methods face challenges in achieving precise nodule segmentation, as many of them depend on inaccurate activation maps or inefficient pseudo-mask generation algorithms.
In this study, we introduce a novel multi-agent reinforcement learning-based WSS framework called Flip Learning, which relies solely on 2D/3D boxes for accurate segmentation. 
Specifically, multiple agents are employed to erase the target from the box to facilitate classification tag flipping, with the erased region serving as the predicted segmentation mask.  
The key contributions of this research are as follows:
1) Adoption of a superpixel/supervoxel-based approach to encode the standardized environment, capturing boundary priors and expediting the learning process. 2) Introduction of three meticulously designed rewards, comprising a classification score reward and two intensity distribution rewards, to steer the agents' erasing process precisely, thereby avoiding both under- and over-segmentation. 3) Implementation of a progressive curriculum learning strategy to enable agents to interact with the environment in a progressively challenging manner, thereby enhancing learning efficiency.
Extensively validated on the large in-house BUS and ABUS datasets, our Flip Learning method outperforms state-of-the-art WSS methods and foundation models, and achieves comparable performance as fully-supervised learning algorithms.
\end{abstract}


\end{frontmatter}

\section{Introduction}
\label{sec:introduction}
\begin{figure*}[!h]
	\centering
	\includegraphics[width=1.0\linewidth]{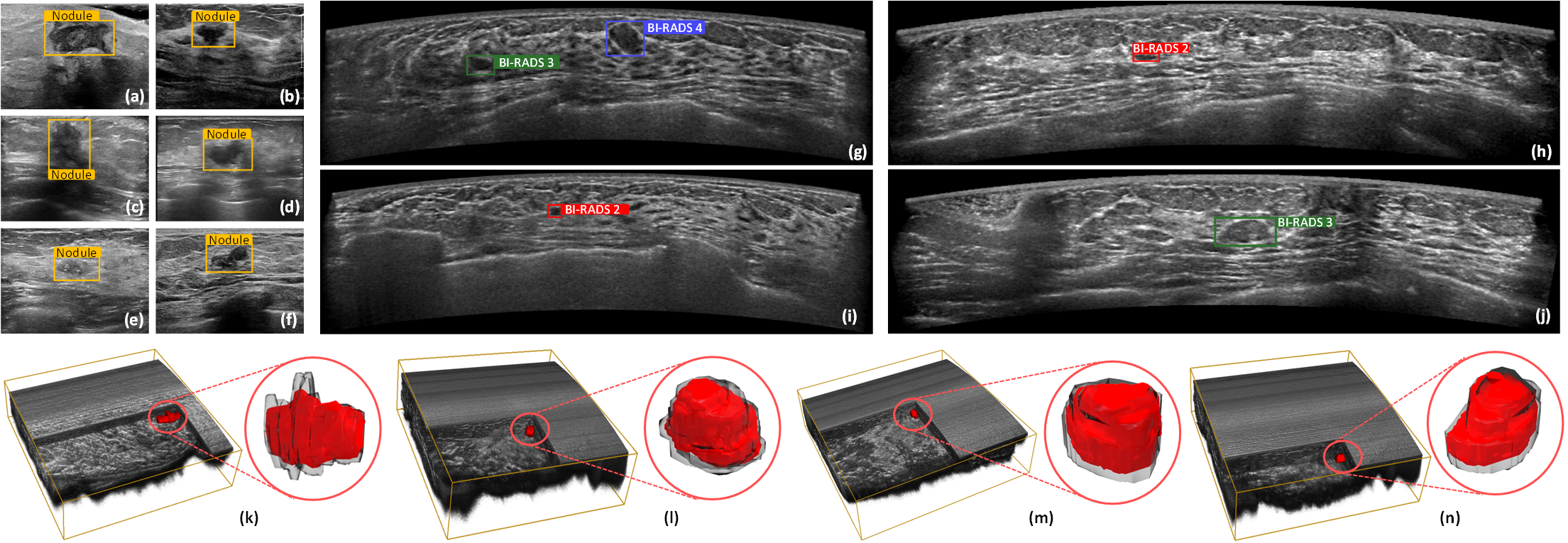}
	\caption{(a)-(f): 2D BUS images with different shapes, sizes and echo characteristics. (g)-(j): Axial views of ABUS images. (k)-(n): 3D visualization of ABUS images and their masks (in the zoom-in red circles, transparent: ground truths, red: predictions). 
 Note that nodules in 2D breast images are labeled without any information about their detailed type. However, the ABUS images are annotated with type tags (i.e., red, green, and blue boxes represent \textit{BI-RADS2-4}, respectively).}
	\label{fig:intro}
\end{figure*}

Breast cancer ranks as the most prevalent cancer among women and stands as a leading cause of cancer-related mortality~\citep{siegel2023cancer}. 
Early identification and treatment of breast nodules are crucial in reducing fatality rates. 
A traditional handheld ultrasound, known as 2D breast ultrasound (BUS), has been recognized as a critical imaging tool in enhancing detection accuracy. It exhibits superior capability in distinguishing dense tissues when compared to mammography~\citep{mandelson2000breast}. Nevertheless, BUS is limited in its capacity to fully encompass and display the entire breast, which affects both user autonomy and reproducibility.

Automated breast ultrasound (ABUS) is a technology that shows promise in addressing these limitations. In contrast to BUS, ABUS follows a standardized acquisition protocol, making it user-friendly for non-experts. Additionally, ABUS can capture an entire breast volume automatically in a single scan, ensuring high reproducibility in breast screening~\citep{wang2019deeply,cao2021dilated,boca2021pros}. Despite the numerous advantages of ABUS imaging, its devices are expensive, which limits its widespread adoption compared to BUS. In conclusion, BUS and ABUS complement each other in various breast screening settings (e.g., community clinics or specialized hospitals). Both modalities have demonstrated effectiveness in accurately detecting nodules at an early stage.

Segmenting the breast nodule from images obtained through BUS and ABUS is a crucial task that can yield valuable clinical information about nodules, such as their shape, size, and boundaries. These clinical parameters play a significant role in assessing risks, diagnosing nodules, and planning treatments. However, manually segmenting these images is labor-intensive and subject to variability, particularly in ABUS images. Examining the entire breast in three-dimensional space consumes a substantial amount of time. Therefore, there is a strong need to develop an automatic and precise segmentation method for both BUS and ABUS images to streamline clinical workflows.

As illustrated in Figure~\ref{fig:intro}, there are several obstacles in automatically segmenting nodules in BUS and ABUS images. Initially, nodules exhibit diverse shapes and sizes due to variations in differentiation and stages, leading to a range of appearance patterns. Some nodules may occupy less than 1\% of the total image volume, making the localization and segmentation of such nodules a challenging and unresolved task. The second difficulty arises from the unclear boundaries of nodules. The distinct intensity distributions of foreground and background, along with their differentiation, can vary significantly in different scenarios, resulting in failures of machine learning algorithms. Lastly, creating pixel-/voxel-level masks for nodules through annotation is laborious and time-intensive; for instance, an ABUS volume may contain over twenty nodules in a single patient. The scarcity of annotated images may impede the development of fully supervised algorithms.

\begin{figure}[!h]
	\centering
	\includegraphics[width=1.0\linewidth]{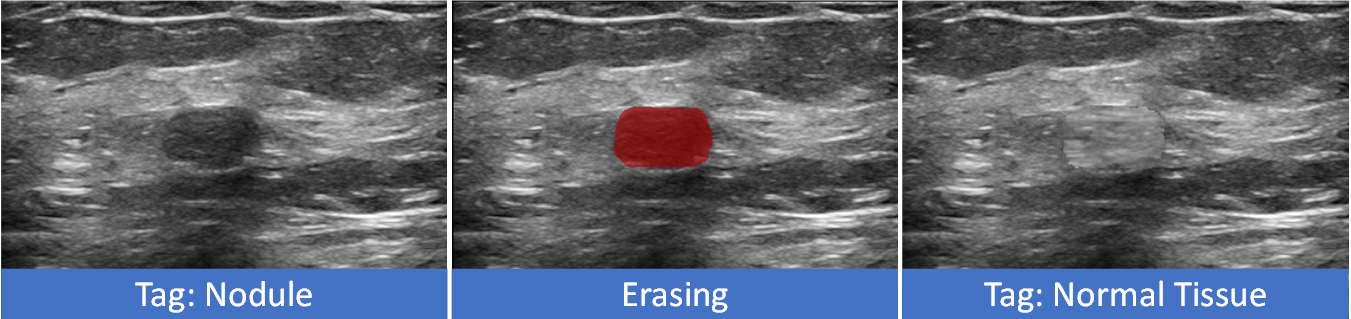}
	\caption{Motivations of flip learning: erasing the nodule from the original image (red region) and inpaint the erased region can flip its tag.
 }
	\label{fig:intro0}
\end{figure}

In this study, we introduce a box-supervised Flip Learning approach for precise segmentation of breast nodules in 2D and 3D US. Our method involves erasing the target from the image to enable flipping the classification tag, as illustrated in Figure~\ref{fig:intro0}. This allows us to obtain segmentation outcomes using the erased area. Our work offers three main contributions:
\begin{itemize}
	\item We view the segmentation task as an erasing task using Multi-agent Reinforcement Learning (MARL).  The environment is represented by superpixels/supervoxels, which allows the utilization of prior boundary information to accelerate the model's learning process. 
    \item We devise three rewards to guide the erasing of agents accurately. Specifically, the classification score reward (CSR) is intended to encourage the erasing to change the image label. The other two intensity distribution rewards (IDR1, IDR2) work to limit the intensity changes in different ways, preventing the issue of excessive segmentation. 
	\item We introduce the strategy of progressive curriculum learning (PCL) to train our proposed framework. By adjusting the number of superpixels or supervoxels, the agents are directed to learn from simple to complex scenarios, enhancing the efficiency of model learning. 
\end{itemize}

\section{Related Works}
\subsection{Supervised Learning for Breast Nodule Segmentation}

Recently, deep learning-based fully supervised segmentation approaches have demonstrated impressive results in medical image segmentation tasks~\citep{liu2019deep,taghanaki2021deep,huang2023test,qureshi2023medical,chang2023pe,yan2024foundation,chang2025p2ed}.
The upcoming section will provide a concise overview of fully supervised research focusing on 2D/3D nodule segmentation in breast US.

In the realm of BUS segmentation, various methods have been suggested to enhance UNet~\citep{ronneberger2015u} with attention mechanisms~\citep{zhang2021sha,punn2022rca,lu2023hybrid,wang2024taichinet}, adversarial learning techniques~\citep{negi2020rda}, multi-scale feature fusion strategies~\citep{xue2021global,shareef2022estan}, and boundary-aware methodologies~\citep{wu2021bgm,xue2021global, huang2022boundary,hu2023boundary}. 
Despite the great performance achieved in 2D BUS segmentation tasks, they may not be suitable to be adopted in the 3D situation directly due to the 1) request for enough computational cost and 2) lack of specific design to capture the 3D spatial information.

In contrast to 2D images, volumetric data contain a greater amount of diagnostic information, making them more effective for clinical analysis. Inspired by the 2D U-net model, the 3D U-net model was introduced to capture diverse features at multiple levels and achieve enhanced segmentation outcomes~\citep{cciccek20163d}.
\citet{wang2019deeply} created a densely deep supervision network along with a threshold loss to segment tumors in ABUS images.
\citet{cao2021dilated} introduced a dilated densely connected U-net that utilized the uncertainty focus loss for segmenting masses in ABUS images.
They then adopted a searchable way to build the auto-densenet for improving the segmentation performance~\citep{cao2022auto}.
Recent studies introduced a cross-model attention mechanism~\citep{zhou2021cross} and a multi-task learning framework~\citep{zhou2021multi} to enhance the segmentation performance.
Most recently,~\citet{pan2023gaussian} introduced a dual decoder design utilizing both CNN and Transformer models. This architecture incorporates an angular margin contrastive loss function to enhance the delineation of breast nodules in ABUS.

However, fully-supervised training requires pixel/voxel-level detailed annotations, which is time-consuming and dependent on expertise. 
Hence, these methods are often constrained by the scarcity of adequate and trustworthy data annotations.

\subsection{Weakly-supervised Learning based Segmentation}
The process of weakly supervised segmentation (WSS) can streamline the annotation pipeline by necessitating only minimal manual annotations (such as image labels, boxes, points, or scribbles) for training the model. In the subsequent section, we primarily focused on examining the methods of WSS at the image and box levels, as these are the most prevalent strategies.

\textbf{Image-level.} 
In the domain of image-level WSS, one of the notable works was Class Activation Mapping (CAM)~\citep{zhou2016learning}. CAM was employed to display the most distinctive characteristics and areas of focus identified by the classifier.
Subsequent research has introduced numerous alterations to the initial CAM method, such as Grad-CAM~\citep{selvaraju2017grad}, Grad-CAM++~\citep{chattopadhay2018grad}, Score-CAM~\citep{wang2020score}, and others.
In the computer vision domain, many WSS techniques were introduced to combine with CAMs to improve the performance, including adversarial erasing~\citep{wei2017object}, affinity labels~\citep{ahn2018learning}, co-attention~\citep{sun2020mining}, sub-class mining~\citep{chang2020weakly}, etc.
There are also lots of related methods for WSS in medical imaging. 
In one of the early explorations,~\citet{gondal2017weakly} proposed to adopt the original CAM with several network architecture modification tricks to achieve good accuracy.
Different from using global average pooling (GAP) in most CNN,~\citet{feng2017discriminative} proposed a multi-GAP strategy to generate nodule activation maps with higher resolution for improving segmentation performance.
~\citet{wu2019weakly} developed a novel CAM using dimensional independent attention for coarsely localizing lesions in 3D space. They also equipped the CAM with a representation model to achieve fine-grained segmentation.
Besides, MS-CAM~\citep{ma2020ms} was proposed to improve the localization performance using multi-level features and attention mechanisms.
\citet{chen2022c} proposed a causal CAM based on two cause-effect chains for medical images.
\citet{feng2023cam} integrated the multiple-instance learning with CAM to alleviate its inaccurate localization problem.
\citet{jiang2024segmentation} introduced a two-stage solution to refine the CAM result with feature fusion and boundary enhancement.
\citet{kuang2024weakly} designed a feature decomposition strategy to solve the challenges in the CAM-based multi-class WSS task.
Moreover, inspired by CAM and anomaly detection, researchers leveraged an anomaly-guided approach based on image-level supervision to obtain pseudo-segmentation labels~\citep{yang2024anomaly}.
However, most image-level WSS methods (i.e., CAM-based) only focus on the most discriminated part of the image, resulting in inaccurate activation maps.
Additionally, they often require sufficient normal and abnormal images, further limiting their clinical availability.

\textbf{Box-level.} Most current box-supervised segmentation approaches highly rely on pseudo-mask generation. 
For example, BoxSup~\citep{dai2015boxsup} and SDI~\citep{khoreva2017simple} highly required the region proposals from MCG~\citep{pont2016multiscale} as labels to train a segmentation network. Similarly, Box2Seg~\citep{kulharia2020box2seg} proposed to generate the mask by GrabCut~\citep{rother2004grabcut} for network learning. 
% Different from the above works, 
DeepCut~\citep{rajchl2016deepcut} used a patch-based classifier to segment brain and lung lesions based on the initial segmentation by GrabCut~\citep{rother2004grabcut}. Recently, BoxInst~\citep{tian2021boxinst} introduced projection and pairwise losses to train the network through the box supervision, without reliance on any pseudo mask generation.
DiscoBox~\citep{lan2021discobox} leveraged box supervision to jointly learn instance segmentation and semantic correspondence.
\citet{li2022box} introduced to integrate level-set evolution with deep learning for effective instance segmentation.
They then extend their work and build Box2Mask~\citep{li2024box2mask} with a stronger ability on WSS.
Moreover, BoxSnake was developed to enhance the box-supervised segmentation with two carefully designed losses~\citep{yang2023boxsnake}.
\citet{xie2023detect} integrated the YOLOV5-based detector into WSS to extract coarse target area via GradCAM.
They then leveraged adaptive region growth to get high-quality pseudo masks.
\citet{wang2024weakly1} equipped the box supervision with the polar transformation-based multiple-instance learning technique to boost the segmentation performance.
Recently,~\citet{wei2024weakpcsod} proposed a mask-to-box transformation and a color consistency loss to achieve box-based WSS for point cloud salient objects.
Though effective in natural image analysis, current box-based approaches will highly rely on ineffective pseudo-label generation algorithms, task-specific loss designs, color-coded differences in foreground and background, etc.
Thus, they may not suit US image segmentation tasks.

\subsection{Weakly-supervised Learning in Breast Nodule Analysis}

This section provides an overview of research that utilizes weakly supervised learning to analyze breast nodules, e.g., classification, segmentation, etc.
\citet{liang2020weakly} presented a method that combines weak supervision with CAM and self-training for segmenting breast tumors in mammography images. 
\citet{shen2021interpretable} suggested a training approach that relies solely on mammography images and image-level annotations (such as indicating the presence of cancer) to generate pixel-level segmentation through the saliency map.
Three CAM-based neural networks were utilized by \citet{kim2021weakly} to identify the cancer location in BUS images.
In the research by \citet{li2022deep}, a segmentor was initially trained to partition the BUS image into four regions: fat, mammary gland, muscle, and thorax.
They subsequently educated a classifier to acquire segmentation outcomes based on CAM for cancer, limited by the specific anatomical details.
Two approaches have been suggested for identifying and categorizing breast tumors by utilizing both the strong Region of Interest (ROI) annotations and weak image-level labels~\citep{shin2018joint,wang2024weakly}.
% Specifically, in~\citep{wang2023weakly}, the authors proposed a two-stage framework to refine coarse ROI-level labels to improve the diagnosis.

\begin{figure*}[!t]
	\centering
	\includegraphics[width=1.0\linewidth]{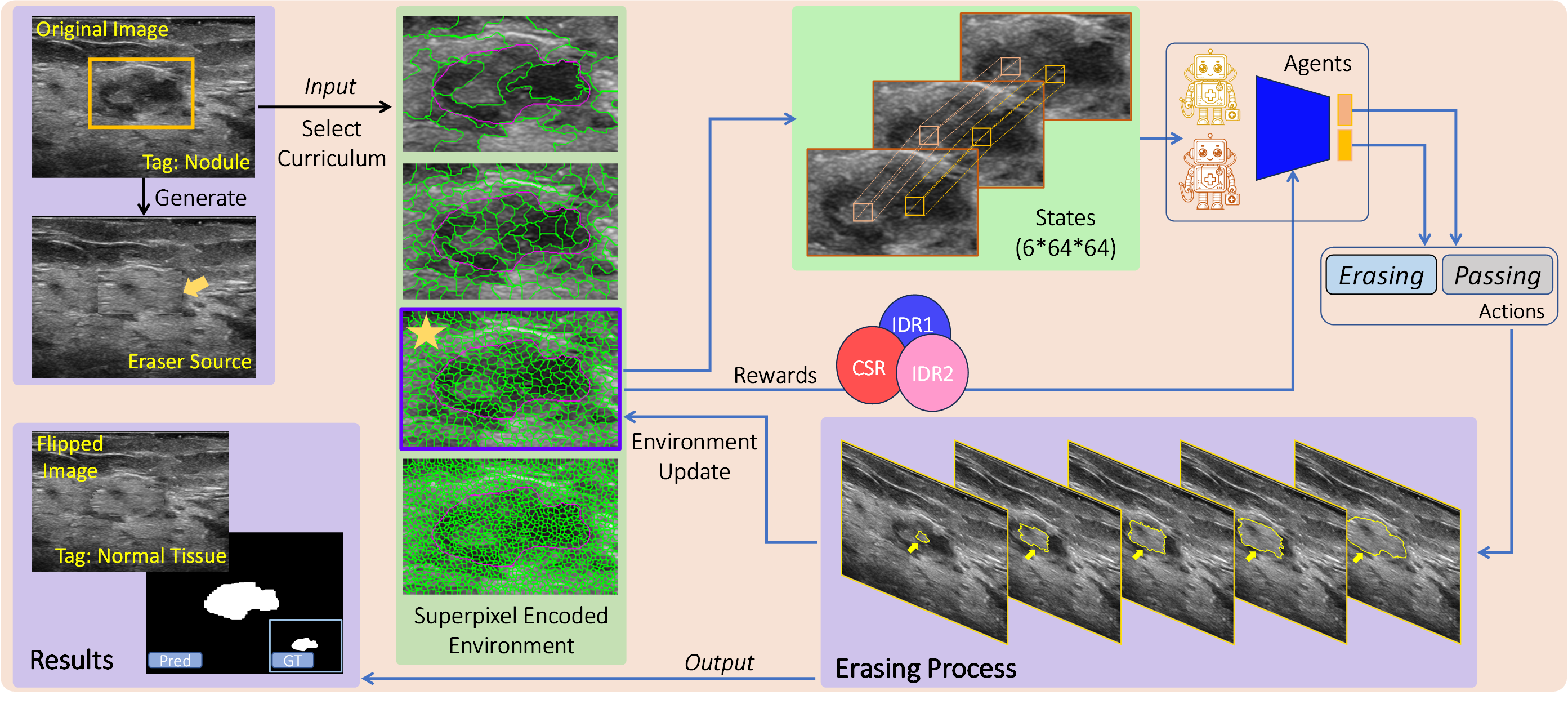}
	\caption{Overview of the proposed framework. Here, we use the BUS image as an example to better illustrate the learning process.
    Purple and green boundaries in the environment block represent the superpixels and annotated masks, respectively.
    The left-upper star shows the environment encoded with the current curriculum setting.
    Yellow boundaries in the erasing process reveal the changes in the erase region.}
	\label{fig:framework}
\end{figure*}

Several research works have focused on employing weakly supervised techniques for the analysis of 3D breast cancer.
In their study, \citet{zhou2019weakly} initially trained a 3D neural network to categorize the BI-RADS of MRI scans. Subsequently, they utilized 3D CAM to pinpoint and outline the cancerous regions, resulting in a satisfactory segmentation DICE score.
In a separate study, \citet{meng2022volume} introduced volume awareness loss and outlier suppression loss to steer the WSS in DCE-MRI using partial annotations at the slice level.
Furthermore, \citet{zhong2023simple} introduced the concept of similarity-aware propagation learning and employed extreme points to supervise the segmentation of cancerous regions in DCE-MRI.

In brief, the current body of research on weakly supervised learning for breast cancer analysis is somewhat limited. Present studies on WSS predominantly depend on CAM, leading to imprecise segmentation because CAM solely identifies the distinctive areas, rather than the precise nodule region. Additionally, there is a lack of weakly supervised research concentrating on 3D breast imaging, specifically ABUS, which impedes the progress of intelligent analysis within the ABUS domain.

\subsection{Reinforcement Learning in Medical Image Analysis}
In a standard reinforcement learning (RL) system, guided by the \textit{reward} signal, an optimal \textit{policy} (i.e., \textit{action} sequences) can be learned through the \textit{agent-environment} interaction. 
Recently, deep RL has been investigated in various domains of medical image analysis~\citep{zhou2021deep,hu2023reinforcement}, e.g., classification~\citep{narmatha2023ovarian}, plane localization~\citep{yang2021agent,yang2021searching,zou2022agent,huang2024localizing}, and segmentation~\citep{bae2019resource,yang2020deep,tian2020multi,zhang2023slide}, demonstrating robust performance and promising prospects.
Nevertheless, the majority of current segmentation methods based on reinforcement learning necessitate manually annotated pixel-level masks for supervised training, which is often impractical.

In our prior MICCAI research, we introduced a MARL-based WSS method called \textit{Flip Learning} to enhance precise segmentation using basic box annotations~\citep{huang2021flip}.
In our proposed framework, the agents erase the nodule within the box, and the erased region is filled based on the pre-generated eraser source. Through erasing, the classification score of \textit{nodule} will drop progressively, making the image tag flip from ``nodule" to ``normal tissue". The erased region will be considered as the final segmentation prediction. 
The motivation for our proposed work can be summarized in mainly two aspects:

\textbf{Why erasing and tag-flipping for segmentation?} Let's say we want to change the tag of an image, for instance, switching from foreground (\textit{nodule}) to background (\textit{normal tissue}). 
See Figure~\ref{fig:intro0}, one of the simplest approaches is to remove the object (i.e., nodule) from the image and then fill the removed area using methods similar to inpainting~\citep{elharrouss2020image}. 
If the erasing is done accurately, it is possible to extract the erased region and convert it into a segmentation mask.

\textbf{What is the rationale behind utilizing RL for representing erasing?} 
RL is centered on maximizing rewards over the long run, making it more appropriate for optimizing long-term goals in tasks involving modeling sequences of actions.
An illustrative task where RL has outperformed conventional supervised learning approaches is \textit{painting}~\citep{huang2019learning,singh2021combining}.
Consequently, employing RL for modeling the process of \textit{erasing} (in contrast to \textit{painting}) is logical and fitting.

In this work, we build upon our MICCAI approach~\citep{huang2021flip} and improve the stability, effectiveness and applicability of \textit{Flip Learning} framework. The variations and enhancements are mainly evident in three crucial areas:
\begin{itemize}
    \item We additionally introduce a reward function to enhance the guidance for the erasing process (IDR2). 
    In contrast to the previous reward (IDR1) that emphasizes the distance between the intensity distributions of erased regions, IDR2 focuses on limiting the gap between the background and foreground, offering supplementary insights for the agents.
    \item We streamline the two-stage coarse-to-fine optimization strategy into a single-stage method utilizing curriculum learning (CL). This adjustment leads to a notable reduction in both training duration and inference time, thereby enhancing the efficiency of the system.
	\item The effectiveness of the proposed framework is demonstrated through experiments on a substantial breast dataset comprising 1) 1388 patients with 2953 BUS images and 2) 602 cases with 1735 ABUS volumes, significantly larger in scale compared to the dataset utilized in our MICCAI study. The results confirm the framework's versatility in segmenting nodules in both 2D and 3D US images.
\end{itemize}

\section{Method}
Figure~\ref{fig:framework} illustrates the MARL-based Flip Learning framework designed for nodule segmentation in BUS and ABUS datasets.
In this framework, multiple agents are deployed to explore the US environment and gradually erase the nodule from the annotated box.
This process involves transforming the ``nodule" into ``normal tissue" as the erasing progresses, leading to tag flipping.
The region that has been erased is considered the final predicted segmentation outcome. 
Initially, the eraser source is determined to guide the filling of the erased area. 
Subsequently, the environment based on the box is encoded using super-pixels/voxels at various levels, following the progressive curriculum settings.
Finally, multiple agents are trained to learn the optimal erasing strategy by leveraging the designed rewards. 

\subsection{Pseudo Samples Generation for Classifier Construction}
In our study, developing a classifier for distinguishing between \textit{nodule} and \textit{normal tissue} is pivotal, serving two main purposes: 1) selecting the eraser source and 2) providing flip signals. For more details, please refer to Sections~\ref{sec:3.2} and~\ref{sec:3.3}.

\begin{figure}[!h]
	\centering
	\includegraphics[width=1.0\linewidth]{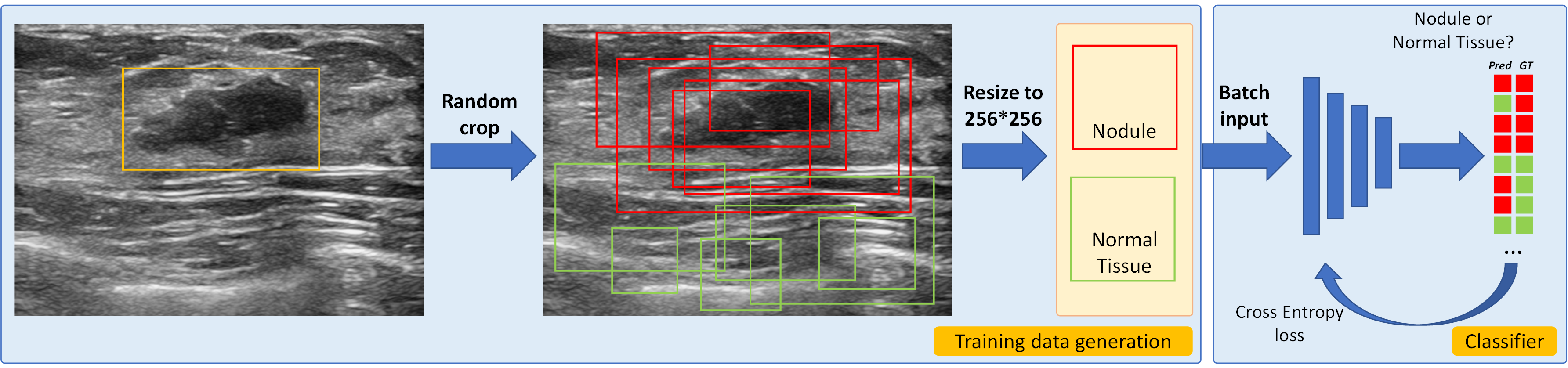}
	\caption{Introduction to the pre-trained classifier.}
	\label{fig:classifier}
\end{figure}

As we only have nodule images from both BUS and ABUS datasets and lack samples from healthy individuals, it is necessary to generate normal samples to facilitate the training of the binary classification network.
To address this, we propose utilizing the self-information of the weak nodule annotation (2D/3D box) to generate adequate normal/abnormal breast data.
In Figure~\ref{fig:classifier}, we extract patches (green boxes) randomly that do not overlap with the nodule annotation (yellow box) to serve as negative data (tag: \textit{normal tissue}). Furthermore, randomly selected boxes that cover more than $50\%$ of the nodule area (red boxes) are identified as positive data (tag: \textit{nodule}) to enhance the diversity of the nodule dataset. These generated samples can then be used to train the classification network.

\subsection{Modeling Patches for Generating Eraser Sources}
\label{sec:3.2}
\begin{figure}[!t]
	\centering
	\includegraphics[width=1.0\linewidth]{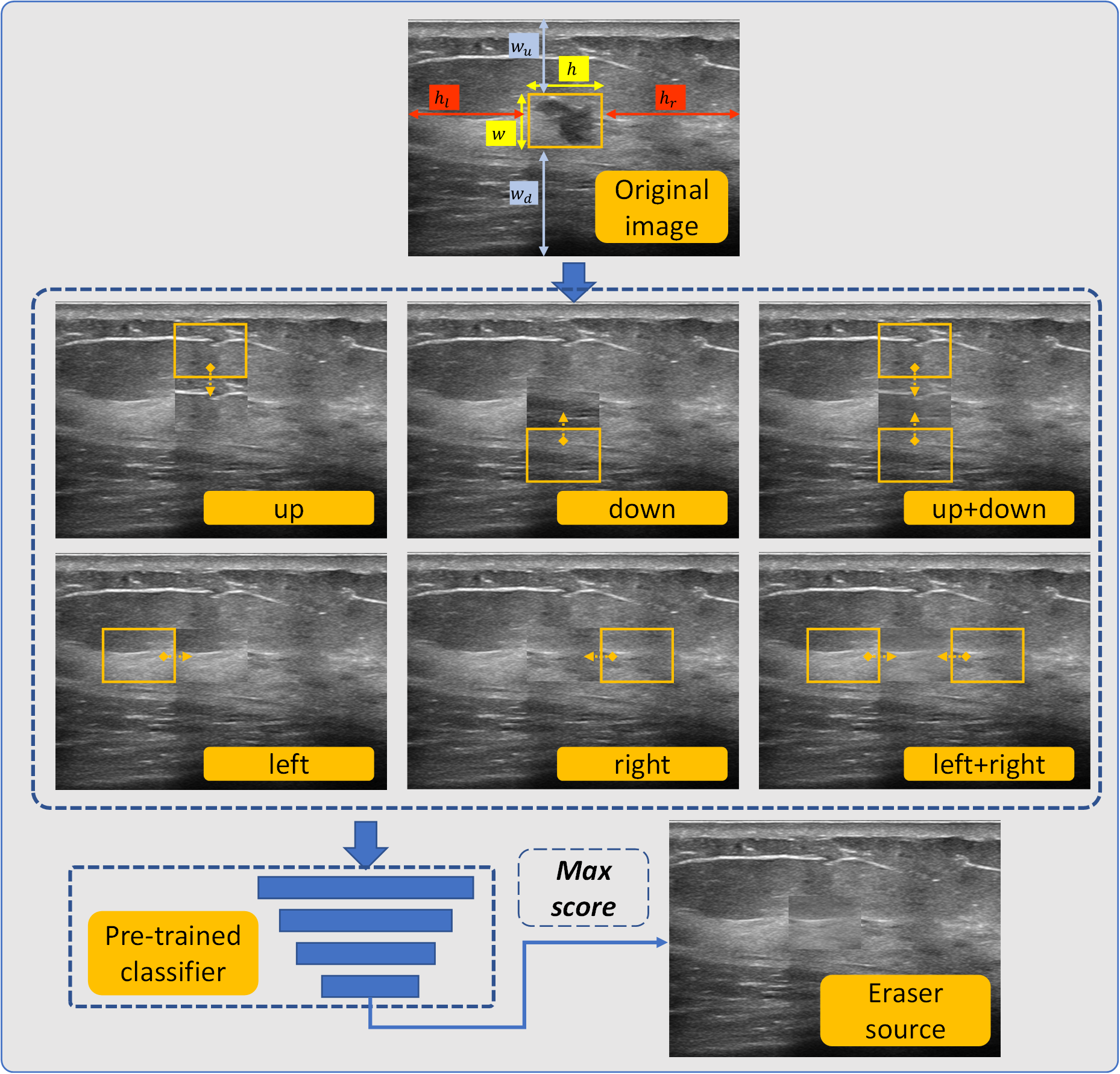}
	\caption{Illustration of eraser source generation.}
	\label{fig:eraser}
\end{figure}

An eraser source can be used to fill the area erased by the agents (see Figure~\ref{fig:framework}). Essentially, the patches surrounding the target area can provide pertinent details to avoid abrupt changes in intensity and content between the target area and the background. As demonstrated in the 2D illustrations in Figure~\ref{fig:eraser}, we opt for a few neighboring fundamental patches (such as left, right, up, down) and also compute their amalgamated version.

\textbf{Creation of primary patches.} 
Assuming that the dimensions (i.e., height and width) of a designated area are $h \times w$ (depicted by the yellow box in the original illustration, Figure~\ref{fig:eraser}), and there is adequate space around the area to extract patches of the same size ($w_u,w_d\ge w$; $h_l,h_r\ge h$), the extracted patches can be directly applied to fill the designated region. In cases where either $(w_u w_d)<w$ or $(h_l h_r)<h$, the patch will be adjusted to $h \times w$ using bi-linear interpolation. Subsequently, the four neighboring patches from perpendicular directions, denoted as $p_u, p_d, p_l, p_r$, can be generated. Expanding this approach to the 3D scenario with the annotated box (length $\times$ width $\times$ height) can be easily achieved by additionally considering the forward and backward patches ($p_f$ and $p_b$).

\textbf{Construction of combined patches.} 
Next, we investigate enhancing the coherence of the content and enhancing the credibility of the eraser source. For the 2D assignment, we examine two approaches to combine the fundamental patches. These are (1) 'up+down' ($p_{ud}$) and (2) 'left+right' ($p_{lf}$), where $p_{ud}=0.5\times(p_u+p_d)$ and $p_{lr}=0.5\times(p_l+p_r)$.
An additional fusion patch can be incorporated into the 3D images by combining the forward and backward patches: $p_{fb}=0.5\times(p_f+p_b)$.
Hence, the BUS dataset comprises 4 primary and 2 combined patches, whereas the ABUS dataset includes 6 primary and 3 combined patches. These patches are transferable to the designated region, creating sets of candidates with 6 and 9 elements for each BUS and ABUS image, respectively. 

In our prior research~\citep{huang2021flip}, we manually identify the optimal eraser source from the candidate pool based on experiential judgment. This process is time-intensive and heavily dependent on the observer's subjective assessment, leading to potentially suboptimal selection and impacting the system's precision.
In this study, we utilize the pre-trained classifier to aid in choosing eraser sources. 
Specifically, as shown in Figure~\ref{fig:eraser}, all candidate images are fed into the classifier, which outputs their probability scores for the category "normal tissue". Subsequently, the image with the highest score is automatically designated as the final eraser source. This approach significantly diminishes the need for operators' subjective judgment and has the potential to identify a more suitable eraser source, thereby enhancing the outcomes of Flip Learning.

\begin{figure*}[!t]
	\centering
	\includegraphics[width=1.0\linewidth]{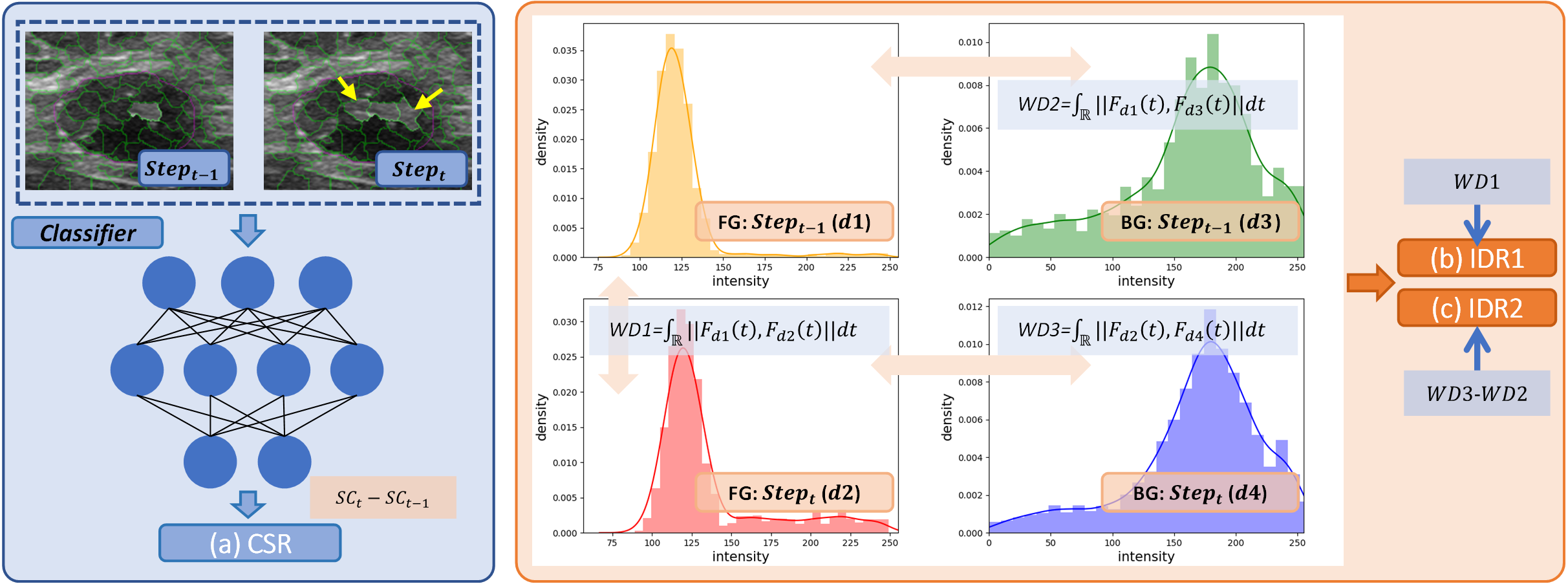}
	\caption{Details of three rewards used in our study, including (a) CSR, (b) IDR1 and (c) IDR2. WD1, WD2, and WD3 calculate the WD distances.}
	\label{fig:reward}
\end{figure*}

\subsection{MARL-based Erasing for WSS}
\label{sec:3.3}
We model the WSS as an erasing task through MARL. 
Within a single RL system, an \textit{agent} explores the \textit{environment} and executes \textit{action} to maximize the accumulated \textit{reward} to acquire the optimal \textit{policy}. 
Ultimately, the interaction between the agent and the environment concludes with the \textit{terminal signal}.
In this study, we suggest employing multiple agents to concurrently explore the environment for increased speed. The fundamental components outlined in MARL are as follows:

-- \textit{Environment:} 
The region where the target (i.e., nodule) is located is referred to as the environment.
Initially, we standardize this environment by resizing it so that the shortest side measures 100 units, while preserving the original aspect ratios of the boxes.
This standardization process can facilitate the learning of agents by maintaining a consistent environment size.
Moreover, when working at the pixel/voxel level, agents are required to interpret dense information without explicit supervised signals. 
To address this, we transform the environment by consolidating pixels/voxels with similar characteristics to create superpixels/supervoxels, as suggested by previous studies~\citep{chen2020review,ibrahim2020applications}. 
The utilization of superpixel/supervoxel techniques has been shown to enhance learning efficiency, reduce computational expenses, and extract valuable shape-related information about the target, thereby offering useful supervised signals.
In this study, we employ the SLIC algorithm~\citep{achanta2012slic} to generate the superpixel/supervoxel for both BUS and ABUS datasets.

-- \textit{Agents:} 
The agent interacts with the environment encoded by superpixels/supervoxels and acquires the most effective erasing strategy. 
To address the extensive search space in 2D/3D US, a multi-agent framework is employed instead of a single agent to enhance efficiency.
Specifically, $\mathcal{K}$ agents collaborate to complete the erasing-based segmentation task.
Following the approach of~\citet{vlontzos2019multiple}, the $\mathcal{K}$ agents are designed to share parameters in the convolution layers to facilitate communication among them.
This sharing of parameters aids in exchanging general knowledge about the environment.
Additionally, $\mathcal{K}$ distinct fully connected layers are integrated after the convolution layers to ensure that each agent comprehensively learns the decision-making process. Each agent is allocated a specific sub-box region (e.g., top and bottom sections in the 2D image) and is tasked with handling a roughly equal number of superpixels/supervoxels (totaling $\mathcal{N}$, with each $\approx\mathcal{N}/\mathcal{K}$).

-- \textit{States:}
The state is typically described as the environment that the agent observes partially at the current step.
Based on the coordinates of one agent, the state is represented by a cropped patch centered on these coordinates, with dimensions of 16$^2$ or 16$^3$ for 2D or 3D scenarios, respectively.
Subsequently, the states of one agent consist of a concatenation of the last three states, encompassing the current state and the two preceding steps.
Selecting three as the number of states to combine strikes a balance between providing ample state information and enhancing learning efficiency~\citep{mnih2015human,yang2021searching}.
In the case of multiple agents (e.g., $\mathcal{K}$), the states can be constructed by aggregating the states from each agent, resulting in a size of \textit{3$\mathcal{K}$}$\times$16$\times$16 or \textit{3$\mathcal{K}$}$\times$16$\times$16$\times$16.

-- \textit{Actions:} 
The agent's action determines whether the current region should be erased or retained.
The action consists of two options: 1) \textit{passing} and 2) \textit{erasing}. 
When the agent chooses the \textit{passing} action, it signifies that the current region is considered background and should not be erased. On the other hand, selecting the \textit{erasing} action means that the superpixel/supervoxel at the current location will be erased and identified as part of the segmentation foreground.
Subsequently, the erased region will be filled based on the generated eraser source.

The agents will navigate through the environment and make decisions to achieve segmentation results through erasing.
The previous traversal approach often requires agents to switch between foreground and background to learn effectively, which may impede their learning process~\citep{huang2021flip}.
Therefore, we have restructured the traversal order to make the agents move roughly from the inner regions toward the outer ones.
We hypothesize that super-regions closer to the center of the box are more likely to be part of the target nodule. By considering the distance between each superpixel/supervoxel and the box center, those with shorter distances will be erased first. This inner-to-outer strategy enables the agents to engage with the target area initially, thereby enhancing their learning stability.

-- \textit{Rewards:} 
The reward function plays a crucial role in RL by assisting agents in acquiring an optimal erasing strategy.
Illustrated in Figure~\ref{fig:reward}, we present a composite reward that comprises three components: (1) CSR, (2) IDR1, and (3) IDR2.
CSR serves as a fundamental element, incentivizing agents to erase the target (i.e., breast nodule) from the box to enhance their classification score and flip the tag from \textit{nodule} to \textit{normal tissue}. Mathematically, it is computed as:
\begin{align}
CSR & = \left\{
\begin{aligned}
+1, & SC_{t}-SC_{t-1}>0 \\
0, & SC_{t}-SC_{t-1} = 0,  \\
-1, & SC_{t}-SC_{t-1}<0
\end{aligned}
\right.
\end{align}
where $SC_{t-1}$ and $SC_{t}$ represent the classification score (\textit{normal tissue}) achieved by the pre-trained classifier at steps \textit{t}-1 and \textit{t}.

However, depending solely on CSR may lead to a significant drawback. 
This approach tends to drive the agents to erase and fill the entire box to achieve a higher score, which could result in excessive erasing and over-segmentation.
Therefore, we introduce IDR1 to impose constraints on the agents' erasing behavior. Specifically, IDR1 limits the discrepancies in intensity distributions between the erased regions at steps \textit{t-1} and \textit{t}.
We utilize the Wasserstein distance (WD) to quantify the disparities between the two distributions. Unlike Kullback-Leibler and Jensen-Shannon divergences, WD is capable of handling arbitrary distributions, even when they do not overlap, and provides meaningful outcomes.
This makes it more suitable for our task, as the distributions involved in the reward computation process may not intersect. IDR1 can be formally defined as:
\begin{align}
	IDR1 & = \left\{
	\begin{aligned}
		+1,& \theta \ge WD1>0 \\
		0,& WD1=0,  \\
		-1,& WD1>\theta  
	\end{aligned}
	\right.
\end{align}
where \textit{WD1} is defined by: $WD(FG_{t-1}, FG_{t})$. A thorough analysis of the $WD$ metric can be found in \citet{villani2009optimal}. 
$FG_{t}$ represents the distribution of the erased region, which is the foreground, at time step $t$.

As previously mentioned, IDR1 specifically limits the intensity distribution within foreground areas. In addition to this, we further encourage agents to increase the separation between the background (non-erased region) and foreground (erased region).
Consequently, the proposed IDR2 can be expressed as:
\begin{align}
IDR2 & = \left\{
\begin{aligned}
+1,& WD3-WD2>0 \\
0,& WD3-WD2=0,  \\
-1,& WD3-WD2<0 
\end{aligned}
\right.
\end{align}
where WD2 and WD3 can be calculated by:
\begin{align}
\label{eq:WD2}
	WD2=WD(BG_{t-1}, FG_{t-1}),
\end{align}
\begin{align}
\label{eq:WD3}
	WD3=WD(BG_{t}, FG_{t}).
\end{align}

The distribution of the non-erased background region at time step $t$ is represented by $BG_{t}$. 
Calculating the IDR2 enables the agents to more effectively understand the connection between the background and foreground, thereby enhancing the accuracy of the erasing process.
The total reward for each agent $Agent_{\mathcal{K}}$ is specified by $R_{\mathcal{K}}=CSR+IDR1+IDR2$

\begin{figure*}[!t]
	\centering
	\includegraphics[width=1.0\linewidth]{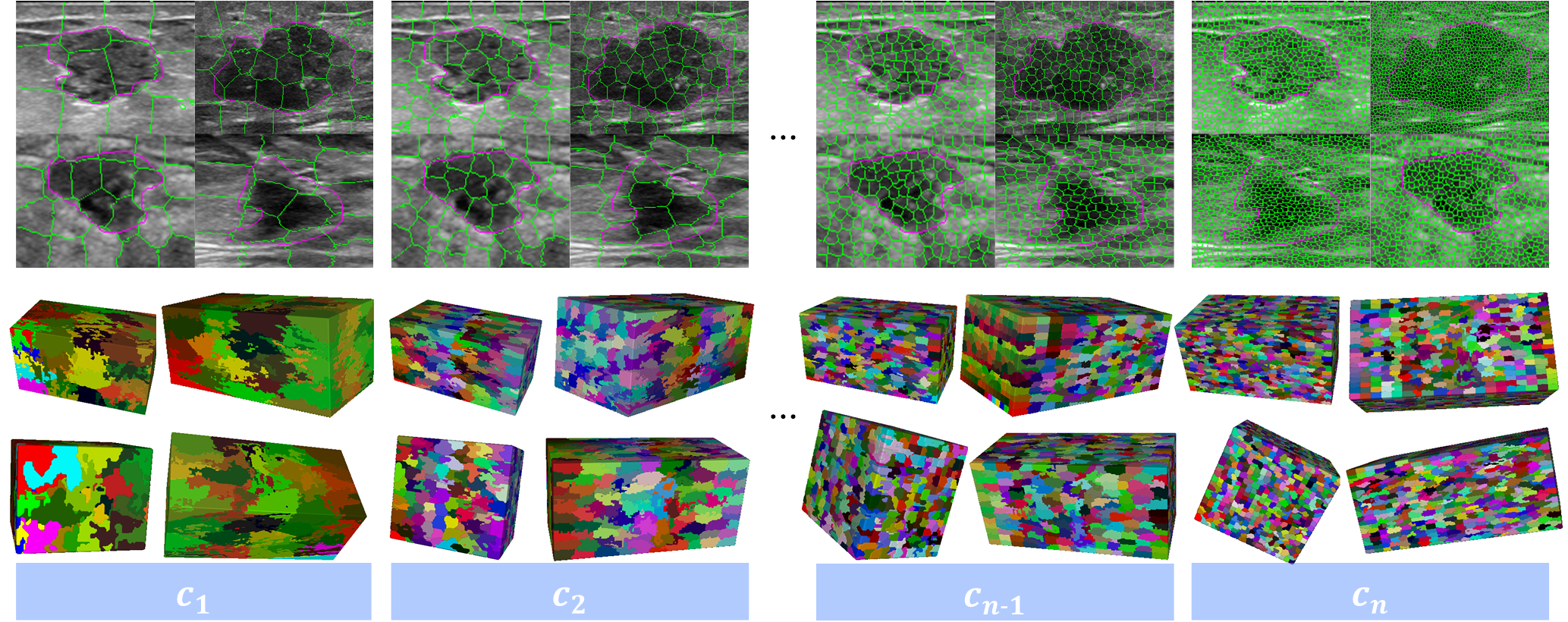}
	\caption{Visualization of the environment with different curriculum settings (from easy to difficult: $c_1$-$c_{n}$).}
	\label{fig:curriculum}
\end{figure*}

-- \textit{Terminal signals:} 
An accurate signal to end the interaction between the agent and the environment is essential.
This research proposes two simple and feasible methods to terminate this interaction.
The initial approach involves setting a limit on the maximum number of times an agent can traverse, denoted as $\mathcal{T}$=2. The second method relies on the classification score, which can intuitively indicate if the erasing process is completed. More precisely, the iteration stops when the nodule's classification score falls below $\mathcal{S}$=0.01.

-- \textit{Algorithms:} 
Deep Q-learning (DQN) techniques are widely used in RL and have been implemented in various medical applications~\citep{yang2021agent,yang2021searching}.
In the vanilla DQN~\citep{mnih2015human}, both action selection and evaluation relied on the \textit{max} operation applied to the target Q-network.
This could lead to Q-value overestimation, resulting in inaccurate model training.
Then,~\citep{van2016deep} introduced the double DQN (DDQN) method, in which separate networks with distinct weights are used for action selection and evaluation (see Eqs.~\ref{eq:choosea} and~\ref{eq:evaluatea}).
They demonstrated that this decoupling can effectively mitigate learning errors caused by inflated Q-values.
\begin{align}
	\label{eq:choosea}
	a^{*} & = \operatorname{argmax}_{\mathrm{a}_{t+1}} \mathrm{Q}\left(\mathrm{s}_{t+1}, \mathrm{a}_{t+1} ; \omega\right),
\end{align}
\begin{align}
	\label{eq:evaluatea}
	\mathrm{Y} & = \mathrm{r}_{t}+\gamma \mathrm{Q}\left(\mathrm{s}_{t+1}, a^{*} ; \omega^{-}\right),
\end{align}
where $\omega$ and $\omega^{-}$ are the parameter weights of current and target \textit{Q} networks, respectively.

Moreover, the sequence data from RL does not meet the criteria of being independently and identically distributed (\textit{i.i.d}) for deep models.
The replay buffer was introduced to be used with DQN, storing the series of states \textit{s}, actions \textit{a}, and rewards \textit{r} (i.e., $\left \{s_t,a_t,r_t,s_{t+1}\right \}$) at each time step \textit{t}.
When training the model, batches of data are selected from the buffer following the uniform distribution, thus disrupting the non-\textit{i.i.d} condition and enhancing the stability of training.
Nevertheless, this method of sampling considers all data points equally important, which could impact the efficiency of learning.
Therefore, \citet{schaul2015prioritized} introduced prioritized experience replay (PER) to enhance the sampling strategy.
In the replay process, data that exhibits larger temporal disparities receive higher weighting.
Then, the training loss of DDQN with PER can be defined as:
\begin{align}
	L & = E_{\left(s_t, a_t, r_t, s_{t+1} \right) \sim M}\left[\left(Y-Q(s_t, a_t ; \omega)\right)^{2}\right]
\end{align}

\subsection{Enhancing Efficiency through Curriculum Learning}
Training an RL system can be challenging and time-consuming because of the absence of supervision signals and the frequent interactions between agents and the environment~\citep{zhou2021deep}.
In our study, we observed that the quantity of superpixels/supervoxels significantly impacts both the efficiency and performance of the model training process.
Specifically, a higher number allows for more precise boundaries, thereby increasing the potential performance ceiling. However, this choice may lead to longer training times and difficulties associated with sparse reward signals.
One potential approach to address this issue is the adoption of a coarse-to-fine (C2F) learning strategy~\citep{tu2008auto,huang2021flip}.
This method involves generating informative signals (coarse masks) in the initial stage and then utilizing them in subsequent stages to enhance learning effectiveness and obtain fine results.

Despite its efficacy, the previous multi-stage solution has two primary drawbacks.
Firstly, the training and testing procedures become complex. Particularly, the reduction in inference speed resulting from the additional learning stages could be unacceptable in clinical reality. Secondly, the practical implementation of multiple models in clinical settings is unfeasible due to the limited memory capacity of US devices. However, in the C2F strategy, distinct learning stages typically necessitate specific models because of variations in input sizes (e.g., feature channels) and learning scopes (e.g., coarse or fine).

In this study, motivated by curriculum learning (CL), we proposed a novel strategy named progressive CL (PCL) to assist the RL model in learning from simple to complex tasks.
The different curricula and their corresponding 2D/3D images are depicted in Figure~\ref{fig:curriculum}.
The complexity of the curriculum is determined by the number of superpixels/supervoxels.
It is important to note that only a single model with a fixed architecture is needed throughout the CL process.
During the training phase, we gradually transition from easy to challenging curricula to encode the environment. 
Specifically, as the training progresses, the number of superpixels/supervoxels in the environment increases incrementally.
PCL enables agents to adapt to the demanding curriculum, capturing prior knowledge of shape and intensity, ultimately achieving satisfactory performance.

\section{Experiments}
\subsection{Clinical Data Collection and Analysis}
We validated our proposed method on two challenging breast datasets, comprising both BUS and ABUS. 
Approved by the local institutional review board (IRB), the BUS images were collected by multiple centers including 8 hospitals (a)-(h).
Besides, the ABUS dataset was obtained from a single center (i).

\textbf{BUS dataset.} 
The specifics of the BUS dataset are outlined in Table~\ref{tab:dataset}.
In our MICCAI 2021 publication~\citep{huang2021flip}, we curated a BUS dataset from Center (a) comprising 1723 images sourced from 1129 patients. In this study, we expanded this dataset by collecting additional data from Center (a), resulting in a total of 3419 images from 1892 patients. 
Subsequently, this dataset was partitioned into subsets of 2401, 338, and 680 images for training, validation, and testing (\textit{Testing-A}) purposes at the patient level.
To further assess the efficacy of our proposed approach, we established a multi-center BUS dataset named \textit{Testing-B} for cross-center validation.
This dataset comprises 325 cases and 1230 images gathered from centers (b)-(h). 
Each image contains a single nodule, which was manually annotated with a mask and a box by sonographers using the Pair annotation software package~\citep{liang2022sketch} (\url{https://www.aipair.com.cn/en/}, Version 2.7).
Annotating times for one mask and box are 2\textit{min} and 6\textit{s}, respectively.

\textbf{ABUS dataset.} 
An overview of the ABUS data is presented in Table~\ref{tab:dataset}.
The dataset comprises 1735 volumes obtained from 605 cases. Each volume contains at least one nodule, with a maximum of 26 nodules. The ABUS dataset was divided randomly into 1124, 125, and 486 volumes at the patient level for training, validation, and testing, respectively.
It is important to highlight that our proposed approach focuses on addressing nodules using 3D boxes, making the nodule count a more precise indicator for model training compared to the number of volumes.
Specifically, the training, validation, and testing sets contain 3073, 318, and 540 nodules, respectively.
The training and validation datasets only include box-level annotations for each nodule, provided manually by experts.
In contrast, the testing set includes both box annotations, voxel-level masks of nodules, and nodule BI-RADS labels, labeled by sonographers under strict quality control.
All the labels were annotated using the Pair annotation software package~\citep{liang2022sketch}.
Annotating one mask and box require 25\textit{min} and 30\textit{s}, respectively.

\begin{table}[htbp]
  \centering
  \scriptsize
  \caption{BUS and ABUS Dataset.}
  \resizebox{0.48\textwidth}{!}{
    \begin{tabular}{cccccc}
    \toprule
          &       & Training & Validation & Testing-A & Testing-B \\
    \midrule
    \multirow{5}[2]{*}{MICCAI} & Case  & 800   & 60    & 269   & / \\
          & Image\&Mask & 1278  & 100   & 345   & / \\
          & Mean Image Size & 320*448 & 320*448 & 320*448 & / \\
          & Mean Nodule Size & 86*133 & 84*125 & 95*141 & / \\
          & Box Area Range & 836-49774 & 1450-30551 & 2156-41538 & / \\
          & Nodule Area Range & 621-34408 & 1193-20115 & 1788-30588 & / \\
    \midrule
    \multirow{5}[2]{*}{BUS} & Case  & 1323  & 189   & 380   & 325 \\
          & Image\&Mask & 2401  & 338   & 680   & 1230 \\
          & Mean Image Size & 320*448 & 320*448 & 320*448 & 727*987 \\
          & Mean Nodule Size & 116*183 & 112*182 & 117*184 & 138*231 \\
          & Box Area Range & 836-138446 & 800-358227 & 1666-122008 & 1763-138788 \\
          & Nodule Area Range & 621-111273 & 579-311604 & 1166-84625 & 1546-107379 \\
    \midrule
          &      & Training & Validation & \multicolumn{2}{c}{Testing} \\
    \midrule
    \multirow{7}[2]{*}{ABUS} & Case  & 385   & 37    & \multicolumn{2}{c}{183} \\
          & Volume & 1124  & 125   & \multicolumn{2}{c}{486} \\
          & Mask/Box & 3073  & 318   & \multicolumn{2}{c}{540} \\
          & Mean Volume Size & 841*226*836 & 839*228*837 & \multicolumn{2}{c}{856*223*838}\\
          & Mean Nodule Size & 56*36*54 & 51*34*50 & \multicolumn{2}{c}{118*68*67} \\
          & Box Area Range & 840-20664644 & 1404-10130505 & \multicolumn{2}{c}{2160-27852264} \\
          & Nodule Area Range & /     & /     & \multicolumn{2}{c}{1498-9874084} \\
          & BI-RADS 2/3/4 & 671/1611/791 & 60/180/78 & \multicolumn{2}{c}{85/220/235} \\
    \bottomrule
    \end{tabular}}%
  \label{tab:dataset}%
\end{table}%

\begin{table*}[htbp]
	\centering
	\scriptsize
	\caption{Traditional and WSS Method comparison on BUS dataset. The best results are shown in bold.}
    \resizebox{0.99\textwidth}{!}{
	\begin{tabular}{ccccccccc}
		\toprule
		& \multicolumn{4}{c}{Testing-A} & \multicolumn{4}{c}{Testing-B} \\
		\midrule
		& DICE$\uparrow$  & JAC$\uparrow$   & HD$\downarrow$    & ASD$\downarrow$   & DICE$\uparrow$  & JAC$\uparrow$   & HD$\downarrow$    & ASD$\downarrow$ \\
		\midrule
        GrabCut~\citep{rother2004grabcut} & 16.41±34.35 & 14.80±31.45 & 13.96±39.11 & 5.11±19.70  & 36.87±44.56    & 33.86±41.36  & 18.74±50.44 & 6.08±24.89   \\
        Saliency-SR~\citep{hou2007saliency} & 49.62±16.34 & 5.17±3.15 & 56.79±23.91 & 34.58±14.66  &  44.53±22.98  &  10.47±10.87 & 65.75±40.09 & 31.63±20.43    \\
        Saliency-FG~\citep{montabone2010human} & 54.11±17.22 & 7.30±6.67 & 55.82±25.52 & 39.11±17.30  &  64.70±25.69  &  11.82±18.60 & 68.72±50.58 & 53.14±28.54    \\
        \midrule
        FullGrad~\citep{srinivas2019full} & 33.97±22.15 & 22.80±17.71 & 89.93±51.39 & 39.55±26.47 & 17.42±18.71 & 10.91±13.51 & 133.58±75.77 & 63.49±39.49 \\
        GradCAM~\citep{selvaraju2017grad} & 35.83±10.04 & 22.27±7.37 & 85.16±45.70 & 40.46±21.20 &
        27.39±16.68 & 17.01±11.82 & 113.65±70.42 & 50.96±32.40 \\
		GradCAM++~\citep{chattopadhay2018grad} & 62.98±12.20 & 47.04±12.19 & 51.69±32.03 & 21.78±13.55 & 60.57±18.96 & 46.01±19.03 & 71.38±47.39 & 25.65±20.41  \\
        EigenGradCAM~\citep{muhammad2020eigen} & 64.35±14.11 & 48.94±14.64 & 55.15±37.29 & 21.62±16.16 &         60.04±17.72 & 45.1±17.53 & 70.68±45.70 & 28.26±20.02  \\
        AblationCAM~\citep{ramaswamy2020ablation} & 65.01±13.64 & 49.59±14.25 & 54.35±37.14 & 21.37±15.86 & 60.39±24.36 & 47.05±22.07 & 74.77±60.30 & 29.95±29.79  \\
        LayerCAM~\citep{jiang2021layerCAM} & 66.40±13.72 & 51.08±13.71 & 49.87±33.22 & 20.09±14.21 & 57.26±20.22 & 42.88±19.73 & 75.63±49.92 & 27.76±22.01  \\
        \midrule
            SDI~\citep{khoreva2017simple} & 64.90±20.47 & 50.92±19.21 &75.81±54.74 & 18.91±14.14 & 33.40±31.77 & 24.82±25.24 & 138.26±152.06 & 44.74±75.50\\
  		BoxInst~\citep{tian2021boxinst} & 87.29±16.68 & 80.02±17.76 & 28.57±28.30 & 5.66±6.94 & 76.58±30.21 & 69.16±29.39 & 65.84±116.05 & 22.77±76.54\\
            DiscoBox~\citep{lan2021discobox} & 87.53±11.63 & 79.23±13.92 & 33.55±29.01 & 5.91±5.25 & 81.64±20.43 & 72.64±21.63 & 71.59±109.87 & 12.69±33.60 \\
            BoxLevelset~\citep{li2022box} & 87.58±12.75 & 79.62±15.40 & 39.15±33.55 & 4.97±6.60 & 78.16±25.81 & 69.54±25.82 & 96.57±150.69 & 10.60±30.43\\
            Box2Mask~\citep{li2024box2mask} & 88.00±12.99 & 80.41±16.01 & 32.88±35.87 & 5.90±8.10 & 83.26±20.09 & 74.94±21.43 & 58.07±82.07 & 14.02±37.31\\
\midrule
         Ours & \textbf{92.37±4.22} & \textbf{86.09±6.80} & \textbf{14.19±13.89} & \textbf{3.69±2.50} & \textbf{92.15±5.61}  & \textbf{85.66±9.36} & \textbf{16.30±17.40}  & \textbf{4.07±4.24} \\
		\bottomrule
	\end{tabular}}%
	\label{tab:bus_comparison}%
\end{table*}%

\begin{table*}[htbp]
  \centering
  \caption{Comparison with supervised-based methods on BUS dataset. The best results are shown in bold.}
  \resizebox{0.99\textwidth}{!}{
    \begin{tabular}{ccccccccc}
    \toprule
          & \multicolumn{4}{c}{Testing-A} & \multicolumn{4}{c}{Testing-B} \\
    \midrule
          & DICE$\uparrow$  & JAC$\uparrow$   & HD$\downarrow$    & ASD$\downarrow$   & DICE$\uparrow$  & JAC$\uparrow$   & HD$\downarrow$    & ASD$\downarrow$ \\
    \midrule
    Unet~\citep{ronneberger2015u}  & 92.97±3.79 & 87.08±6.25 & 23.30±19.12 & 3.63±2.23 & 90.49±4.60 & 82.93±7.32 & 27.16±24.01 & 5.87±3.98 \\
    nnU-Net~\citep{isensee2021nnu} & \textbf{94.17±3.56} & \textbf{88.15±6.31} & 18.17±15.17 & \textbf{3.34±2.17} & 91.55±3.67 & 84.17±7.03 & 26.53±23.83 & 5.62±4.14 \\
    Transunet~\citep{chen2021transunet} & 91.15±4.56 & 85.17±7.14 & 25.45±20.83 & 4.67±2.61 & 88.71±4.13 & 81.43±8.22 & 29.85±26.72 & 6.14±5.00 \\
    Swin-unet~\citep{cao2022swin} & 93.44±4.01 & 87.56±6.43 & 21.47±17.89 & 3.72±2.77 & 90.99±4.52 & 83.78±7.63 & 26.94±23.87 & 5.49±4.28 \\
    U-mamba~\citep{ma2024u} & 91.87±4.24 & 85.92±6.91 & 23.55±18.91 & 3.90±2.55 & 89.27±5.21 & 80.01±8.92 & 30.54±25.77 & 6.20±5.22 \\
    Swin-Umamba~\citep{liu2024swin} & 92.18±4.30 & 85.89±6.78 & 23.17±18.78 & 3.72±2.43 & 90.44±4.20 & 81.19±9.14 & 27.05±24.17 & 5.62±4.39 \\
    \midrule
    Ours  & 92.37±4.22 & 86.09±6.80 & \textbf{14.19±13.89} & 3.69±2.50 & \textbf{92.15±5.61} & \textbf{85.66±9.36} & \textbf{16.30±17.40} & \textbf{4.07±4.24} \\
    \bottomrule
    \end{tabular}%
  \label{tab:sup-bus}}%
\end{table*}%

\begin{table*}[!t]
  \centering
  \footnotesize
  \caption{Annotation cost (AC) and semi-supervised analysis for different methods. The best results are shown in bold.}
  \setlength{\tabcolsep}{5.2mm}{
    \begin{tabular}{ccccccc}
    \toprule
          &       & Anno. Num & DICE$\uparrow$  & JAC$\uparrow$   & HD$\downarrow$    & ASD$\downarrow$ \\
    \midrule
    \multirow{7}[6]{*}{AC=1} & Unet  & 30    & 65.18±10.78 & 52.77±12.33 & 44.35±24.77 & 18.71±15.67 \\
          & nnU-Net & 30    & 70.44±9.22 & 61.49±11.86 & 40.56±20.13 & 15.31±13.99 \\
          & Ours  & 360   & 80.67±8.10 & 72.84±9.66 & 20.17±14.33 & 6.23±5.14 \\
\cmidrule{2-7}          & (Semi) Unet & 30+2371 & 66.72±9.53 & 53.08±10.76 & 42.98±24.10 & 17.96±14.17 \\
          & (Semi) Unet+Ours & 30+2371 & 80.48±6.74 & 71.26±8.17 & 24.65±16.11 & 6.21±4.76 \\
\cmidrule{2-7}          & (Semi) nnU-Net & 30+2371 & 72.96±8.74 & 62.79±10.64 & 37.62±20.03 & 13.10±10.14 \\
          & (Semi) nnU-Net+Ours & 30+2371 & 84.66±5.16 & 77.43±7.33 & 20.12±17.34 & 5.66±4.73 \\
    \midrule
    \multirow{7}[6]{*}{AC=5} & Unet  & 150   & 71.78±9.83 & 60.54±10.17 & 36.33±21.29 & 16.00±12.63 \\
          & nnU-Net & 150   & 76.74±7.31 & 68.21±9.08 & 28.14±17.62 & 11.18±9.71 \\
          & Ours  & 1800  & 87.59±6.03 & 78.87±8.32 & 18.22±14.45 & 6.19±5.21 \\
\cmidrule{2-7}          & (Semi) Unet & 150+2251 & 73.56±10.99 & 61.87±12.86 & 35.49±22.18 & 15.66±13.74 \\
          & (Semi) Unet+Ours & 150+2251 & 87.44±6.72 & 78.01±9.12 & 23.06±18.33 & 6.28±4.98 \\
\cmidrule{2-7}          & (Semi) nnU-Net & 150+2251 & 78.92±8.45 & 69.18±10.17 & 26.52±18.64 & 11.53±10.02 \\
          & (Semi) nnU-Net+Ours & 150+2251 & 93.88±3.27 & \textbf{89.01±6.52} & \textbf{17.68±15.24} & 3.41±2.38 \\
    \midrule
    \multirow{2}[2]{*}{AC=80} & Unet  & 2401  & 92.97±3.79 & 87.08±6.25 & 23.30±19.12 & 3.63±2.23 \\
          & nnU-Net & 2401  & \textbf{94.17±3.56} & 88.15±6.31 & 18.17±15.17 & \textbf{3.34±2.17} \\
    \bottomrule
    \end{tabular}}%
  \label{tab:time_semi}%
\end{table*}%

\begin{table*}[!t]
  \centering
  \caption{Comparison with foundation models on BUS dataset. The best results are shown in bold.}
  \resizebox{0.99\textwidth}{!}{
    \begin{tabular}{cccccccccc}
    \toprule
          &       & \multicolumn{4}{c}{Testing-A} & \multicolumn{4}{c}{Testing-B} \\
    \midrule
          &       & DICE$\uparrow$  & JAC$\uparrow$   & HD$\downarrow$    & ASD$\downarrow$   & DICE$\uparrow$  & JAC$\uparrow$   & HD$\downarrow$    & ASD$\downarrow$ \\
    \midrule
    \multirow{4}[2]{*}{Tight box} & SAM-Ori~\citep{kirillov2023segment} & 91.26±4.71 & 84.24±7.45 & 22.21±18.86 & 4.25±2.40 & 91.97±3.94 & 85.86±6.48 & 22.66±21.62 & 5.06±3.93 \\
          & SAM-FT~\citep{huang2024segment} & 92.30±5.00 & 85.99±7.02 & 20.18±17.96 & 4.11±2.36 & \textbf{92.99±4.58} & \textbf{86.71±6.57} & 21.17±20.58 & 5.00±4.07 \\
          & MedSAM~\citep{ma2024segment} & 92.35±4.82 & \textbf{86.21±6.67} & 20.33±17.89 & 4.20±2.61 & 92.01±6.20 & 85.08±7.44 & 23.02±21.11 & 5.11±4.62 \\
          & Ours  & \textbf{92.37±4.22} & 86.09±6.80 & \textbf{14.19±13.89} & \textbf{3.69±2.50} & 92.15±5.61 & 85.66±9.36 & \textbf{16.30±17.40} & \textbf{4.07±4.24} \\
    \midrule
    \multirow{5}[2]{*}{0-10 pixels} & SAM-Ori~\citep{kirillov2023segment} & 87.38±6.45 & 81.45±8.23 & 26.77±20.08 & 5.74±3.89 & 86.24±4.55 & 80.70±7.02 & 24.78±21.44 & 5.99±4.23 \\
          & SAM-FT~\citep{huang2024segment} & 89.26±6.31 & 82.05±9.11 & 25.89±21.02 & 5.49±3.62 & 88.05±5.10 & 82.17±6.90 & 22.83±21.14 & 5.31±4.02 \\
          & MedSAM~\citep{ma2024segment} & 89.14±5.77 & 81.65±9.00 & 26.02±20.47 & 5.63±4.04 & 87.54±6.10 & 82.01±7.71 & 23.35±20.27 & 5.66±4.61 \\
          & RoBox-SAM~\citep{huang2024robust}& 90.78±5.36 & 83.32±6.98 & 22.18±19.61 & 4.96±3.66 & 89.22±5.90 & 82.99±6.19 & 21.66±19.74 & 5.00±5.19 \\
          & Ours  & \textbf{91.08±5.23} & \textbf{85.78±7.04} & \textbf{15.21±14.17} & \textbf{3.79±2.46} & \textbf{90.87±4.87} & \textbf{85.34±9.07} & \textbf{18.05±18.11} & \textbf{4.27±4.10} \\
    \bottomrule
    \end{tabular}}%
  \label{tab:SAM-WSS}%
\end{table*}%

\begin{figure*}[!h]
	\centering
 \includegraphics[width=1.0\linewidth]{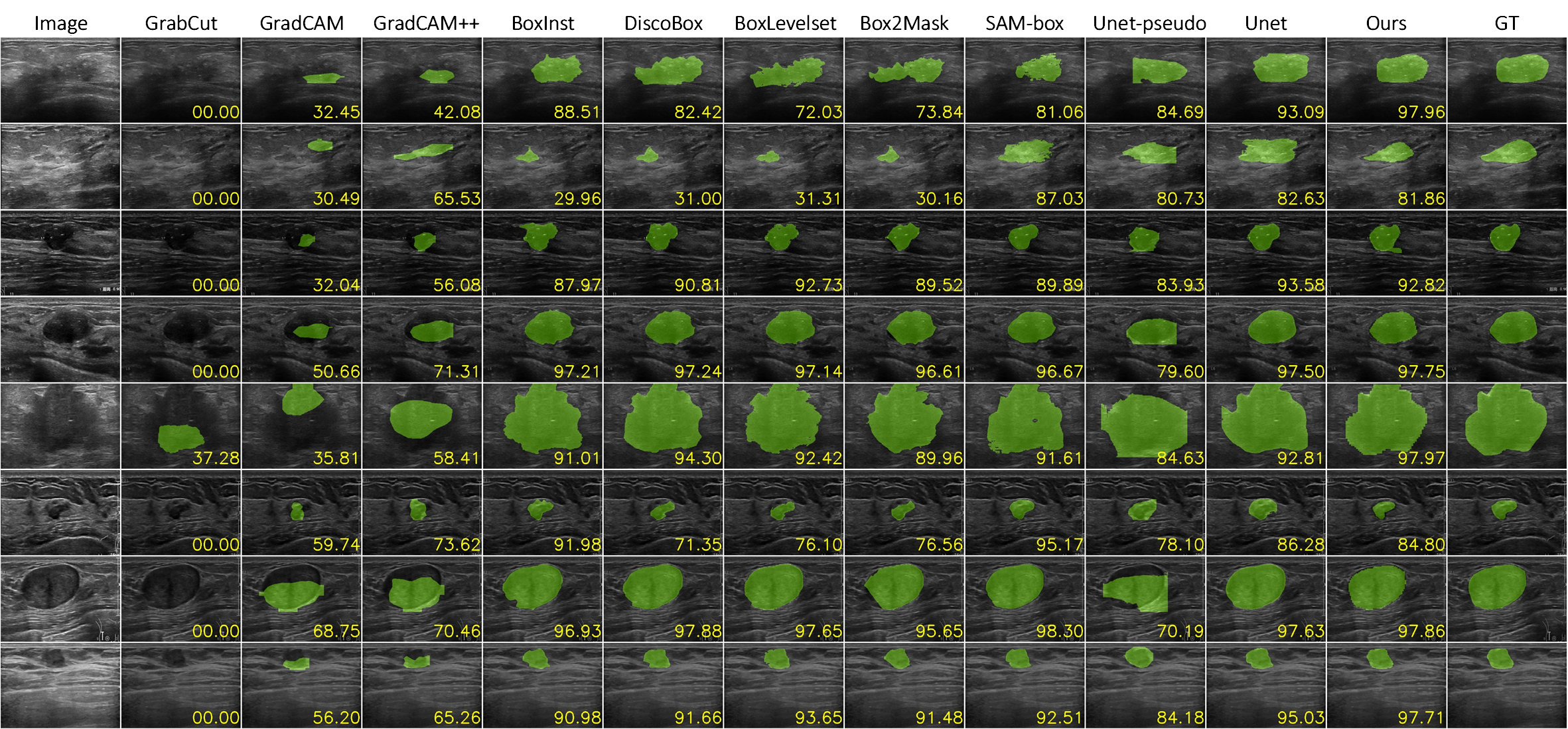}
	\caption{BUS results of different methods, including traditional method (GrabCut), CAM- and box-based WSS methods, SAM, Unet-based supervised methods and Ours. Rows 1-5 are images from the Testing-A set, and the last 3 rows are images from the Testing-B set.}
	\label{fig:result_2d}
\end{figure*}

\begin{figure*}[!h]
	\centering
	\includegraphics[width=0.99\linewidth]{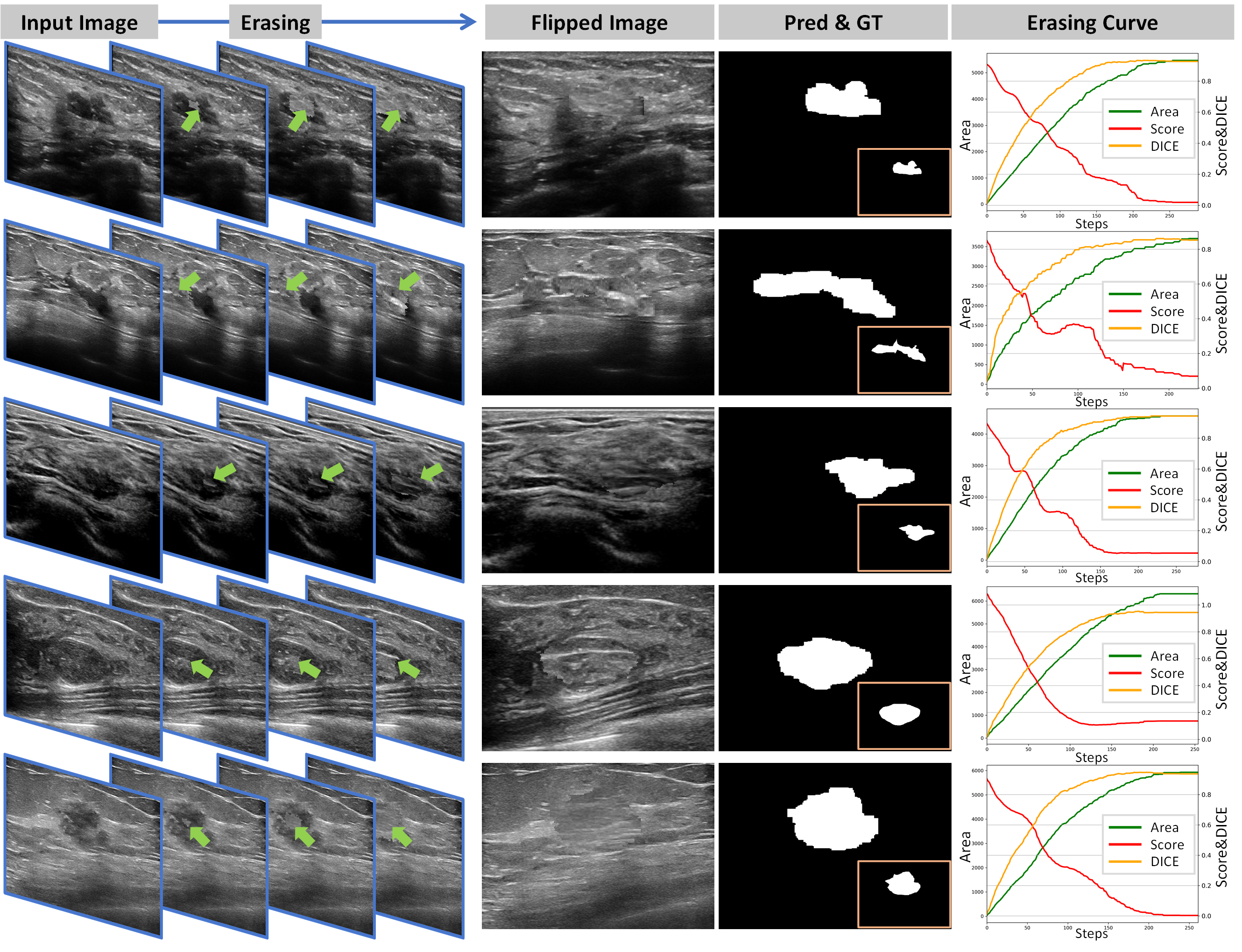}
	\caption{Five typical cases with good BUS nodule segmentation results. For each case (row), columns 1-4 visualize the erasing process in different steps. Columns 5 and 6 show the final predicted masks and GTs. The last column visualizes the erasing curves.}
	\label{fig:2D_result_step}
\end{figure*}

\subsection{Experimental Setup and Evaluation Metrics}
In this study, we implemented the proposed framework in Pytorch, using an NVIDIA TITAN 2080 GPU with 12G memory.
The classification networks employed 2D and 3D ResNet18 as backbones.
AdamW optimizers were used for both networks, with learning rates set to 1e-3.
The batch sizes were 128 for the BUS classifier and 16 for the ABUS classifier.
Data augmentation was performed online, incorporating standard augmentation techniques such as random rotation, shifting, and translation, along with the superpixel/supervoxel-based filling approach.
Specifically, we utilized the eraser source to fill the original image at the superpixel/supervoxel level, and the filling area ratio ($r$) was adjusted to determine the type of augmented images.
Images with $r \leq 0.5$ were classified as \textit{nodule}, whereas those with $r \geq 0.5$ were categorized as \textit{normal tissue}.
The classification performance (accuracy, F1-score, etc.) of different models can be found in the supplementary materials.
During the RL training phase, the agents were also constructed using 2D/3D ResNet18 networks.
They underwent training using the AdamW optimizer for 100 and 150 epochs, respectively, with a learning rate of 5e-5.
The replay buffers had sizes of 8000 and 2000 for 2D and 3D images. The target Q-networks were updated by copying the parameters of the current Q-networks every 1200 iterations for both cases.
The value of $\theta$ in the IDR1 calculation was set to 25.
To generate superpixels/supervoxels, we utilized the $Scikit$-$image$ Python package, primarily based on the hyper-parameter $n_{segment}$.
The curriculum difficulty ($c_n$) was increased every 20 epochs. For BUS images, we defined $c_1$-$c_3$, with corresponding curricula (i.e., $n_{segment}$) set to 100, 1000, and 2000. Additionally, we implemented a five-level curriculum for ABUS images, with $n_{segment}$ values of 100, 1000, 2000, 5000, and 10000.
Training 2D BUS data takes approximately 2 days using a batch size of 64, whereas training the ABUS data with a batch size of 16 takes around 4 days.

The segmentation performance was evaluated using four metrics: DICE similarity coefficient (DICE-\%), Jaccard index (JAC-\%), Hausdorff distance (HD-pixel/voxel), and Average surface distance (ASD-pixel/voxel). DICE and JAC focus on determining segmentation accuracy by quantifying the agreement between the prediction and ground truth (GT), whereas HD and ASD gauge geometric accuracy by assessing boundary distances. Therefore, these metrics offer an objective and comprehensive assessment of segmentation performance.

\subsection{Traditional and WSS Method Comparison on BUS Dataset}
Table~\ref{tab:bus_comparison} shows the BUS results performed by different methods, including the 
\textbf{three} traditional algorithms (GradCut, Saliency-SR and Saliency-FG), \textbf{six} CAM-based WSS methods (EigenGradCAM, FullGrad, LayerCAM, GradCAM, GradCAM++, and AblationCAM), and
\textbf{five} box-based WSS methods (SDI, Boxinst, Discobox, Boxlevelset and Box2mask).

In \textbf{Testing-A},
conventional methods like \textit{GradCut} and \textit{Saliency}-based techniques exhibit subpar results in the BUS segmentation task, possibly due to their strong reliance on RGB color representation, which is incompatible with the grey-scale nature of BUS images.
Most CAM-based WSS methods demonstrate suboptimal accuracy, with the most effective one achieving only a 66.40\% DICE score.
Among WSS methods driven by boxes, the majority (4/5) surpasses a DICE metric of 87\%, showing satisfactory performance.
See the last line, \textit{Ours} demonstrates superior performance across all evaluation metrics compared to all other methods.

We also present the findings of \textbf{Testing-B} in Table~\ref{tab:bus_comparison}. 
\textit{Testing-B} represents a multi-center dataset that is distinct from the training, validation, and \textit{Testing-A} datasets.
The use of different domains for training and testing may lead to a decrease in performance for deep learning models~\citep{huang2022online} (the intensity distribution shift between two datasets can be found in supplementary materials).
Hence, the performance on \textit{Testing-B} can effectively demonstrate the models' generalization capability, as a robust model should exhibit consistent performance across various test sets.
The table reveals that most of the methods experience significant performance declines.
For instance, SDI's DICE metric drops from 64.90\% to 33.40\% (31.50\%$\downarrow$) across the two testing sets.
This indicates the limited generalization ability of previous WSS methods. 
The final row demonstrates that \textit{Ours} only experiences a slight 0.22\% reduction in the DICE score in \textit{Testing-B} compared to \textit{Testing-A} set, highlighting the strong generalization ability of our \textit{Flip Learning} approach (detailed analysis refers to the Sec.~\ref{sec:ablation}).

We additionally test our model on three public datasets, including BrEaST~\citep{pawlowska2024curated}, STU~\citep{zhuang2019rdau} and BUSI~\citep{al2020dataset} (benign and malignant).
Dataset details are provided in the supplementary materials.
Our method achieves satisfactory performance on these external datasets, and the average DICE metrics for BrEaST, STU, BUSI-benign and BUSI-malignant are 90.97\%, 91.20\%, 91.94\% and 89.48\% respectively.
The full results are available in the supplementary materials.
This further proves the strong generalization performance of our approach.

\begin{table}[!t]
	\centering
        \scriptsize
	\caption{Fully-supervised method comparison on different ABUS datasets.}
        \resizebox{0.48\textwidth}{!}{
	\begin{tabular}{cccc}
		\toprule
		& Case/Volume/Target & DICE$\uparrow$  & JAC$\uparrow$ \\
		\midrule
		\citet{tan2016segmentation} & 64/75/78 & $0.73_{0.14}$ & * \\
        \citet{kozegar2017mass} & 32/42/50 & $0.74_{0.19}$ & *\\
		\citet{agarwal2018lesion} & 28/56/56 & $0.69_{0.11}$ & * \\
		\citet{wang2019deeply} & 219/614/745 & $0.58_{0.26}$ & *\\
		\citet{cao2021dilated} & 107/170/170 & $0.69_{*}$ &  *\\
        \citet{zhou2021multi} & 107/170/170 & $0.78_{0.15}$ & $0.65_{0.17}$ \\
		\citet{cao2022auto} & 107/170/170 & $0.78_{*}$ & $0.64_{*}$ \\
        \citet{pan2023gaussian} & 107/170/170 & $0.81_{*}$ & * \\
		\bottomrule
	\end{tabular}}%
	\label{tab:abus_comparison}%
\end{table}%

\begin{table*}[!t]
	\centering
        \scriptsize
	\caption{Weakly-supervised Method comparison on ABUS dataset. The best results are shown in bold.}
    \resizebox{0.9\textwidth}{!}{
	\begin{tabular}{cccccc}
		\toprule
		& Methods & DICE$\uparrow$  & JAC$\uparrow$   & HD$\downarrow$    & ASD$\downarrow$ \\
		\midrule
		\multirow{14}[2]{*}{2D to 3D} 
        & GradCAM~\citep{selvaraju2017grad} & 25.70±17.49 & 15.92±11.82 & 32.56±10.68 & 9.54±2.97 \\
		& GradCAM++~\citep{chattopadhay2018grad} & 25.53±17.35 & 15.79±11.69 & 32.59±10.60 & 9.60±3.04 \\
		& AblationCAM~\citep{ramaswamy2020ablation} & 23.49±16.84 & 14.37±11.21 & 32.67±9.98 & 10.41±3.72 \\
        & EigenGradCAM~\citep{jacobgilpytorchCAM}& 25.78±17.52 & 15.98±11.85 & 32.55±10.72 & 9.52±2.93 \\
		& LayerCAM~\citep{jiang2021layerCAM} & 25.18±17.25 & 15.54±11.58 & 32.6±10.48 & 9.71±3.15 \\
		& FullGrad~\citep{srinivas2019full} & 27.59±19.33 & 17.52±13.57 & 32.65±10.77 & 9.46±3.52 \\
            \cline{2-6}
            & SDI~\citep{khoreva2017simple} & 42.15±17.82 & 28.15±15.09 & 40.82±20.19&13.81±21.91\\
            & BoxInst~\citep{tian2021boxinst} &51.63±10.27& 42.67±9.13 & 25.87±12.11 & 5.72±6.11\\
            & DiscoBox~\citep{lan2021discobox} & 52.24±9.28 & 43.98±10.17 & 24.09±11.87 & 6.05±7.01\\
            & BoxLevelset~\citep{li2022box} & 52.88±11.52 & 43.87±10.01 & 24.67±12.09 & 5.98±6.46\\
            & Box2Mask~\citep{li2024box2mask} & 52.77±9.77 & 43.21±9.87 & 24.98±10.72 & 6.24±7.22\\
            \cline{2-6}
            & SAM-everything~\citep{kirillov2023segment} & 8.55±8.86 & 4.74±6.06 & 497.73±99.80 & 191.18±47.19 \\ %1.93±2.27
            & SAM-point~\citep{kirillov2023segment} & 37.99±26.27 & 27.13±22.89 & 388.57±189.03 & 106.40±73.94 \\ %1.15±0.87 
            & SAM-box~\citep{kirillov2023segment} & 65.36±15.00 & 50.23±15.30 & 41.69±21.53 & 5.52±4.91 \\
            & SAM-Med-box~\citep{huang2024segment}  & 66.17±17.81 & 51.46±16.14 & 40.17±20.99 & 6.05±5.81 \\
          & MedSAM-box~\citep{ma2024segment} & 66.01±16.14 & 51.17±16.32 & 38.76±20.10 & 5.77±4.65 \\
		\midrule
        \multirow{6}[3]{*}{Video} 
        & SAM2-F~\citep{ravi2024sam} & 60.14±20.14 & 45.23±16.08 & 44.17±22.56 & 7.98±5.05 \\
        & SAM2-M~\citep{ravi2024sam} & 65.77±16.25 & 51.99±17.05 & 43.38±21.16 & 7.65±4.71 \\ 
        & MedSAM2-F~\citep{zhu2024medical} & 62.24±18.33 & 47.56±15.86 & 45.86±22.05 & 7.83±4.92 \\
        & MedSAM2-M~\citep{zhu2024medical} & 68.30±14.47 & 53.78±14.91 & 42.99±21.17 & 6.23±4.24 \\
        & SAM2-MedIV-F~\citep{ma2024segment2} & 63.08±17.74 & 48.21±15.74 & 44.32±20.03 & 7.40±5.13 \\
        & SAM2-MedIV-M~\citep{ma2024segment2} & 68.97±14.60 & 54.48±13.97 & 41.80±19.94 & 6.11±4.56 \\ 
        \midrule
		\multirow{7}[1]{*}{3D} 
		& GradCAM~\citep{selvaraju2017grad} 
        & 40.77±8.01 & 25.92±6.34 & 31.01±4.34 & 6.02±1.04 \\
		& GradCAM++~\citep{chattopadhay2018grad}  & 47.63±9.10 & 31.72±7.79 & 26.95±4.63 & 5.83±1.12 \\
        & SAM-Med3D-p1~\citep{wang2024sam} & 38.44±18.18 & 27.39±14.33 &42.78±18.80 & 8.82±6.91\\
        & SAM-Med3D-p3~\citep{wang2024sam} & 43.27±18.58 & 29.39±37.18 & 37.18±18.56 & 7.77±6.62\\
        & SAM-Med3D-p5~\citep{wang2024sam} & 45.77±18.32 & 31.46±15.18 & 36.58±17.25 & 7.61±5.94\\
        & SAM-Med3D-p10~\citep{wang2024sam} & 48.03±18.53 & 33.51±15.73 & 35.72±16.92 & 7.40±6.29\\
		\midrule
		3D & Ours  & \textbf{75.48±12.67} & \textbf{61.22±10.41} & 
		\textbf{20.08±18.16} &\textbf{3.09±1.83} \\
		\bottomrule
	\end{tabular}}%
	\label{tab:abusweak}%
\end{table*}%

\begin{figure*}[!t]
	\centering
	\includegraphics[width=1.0\linewidth]{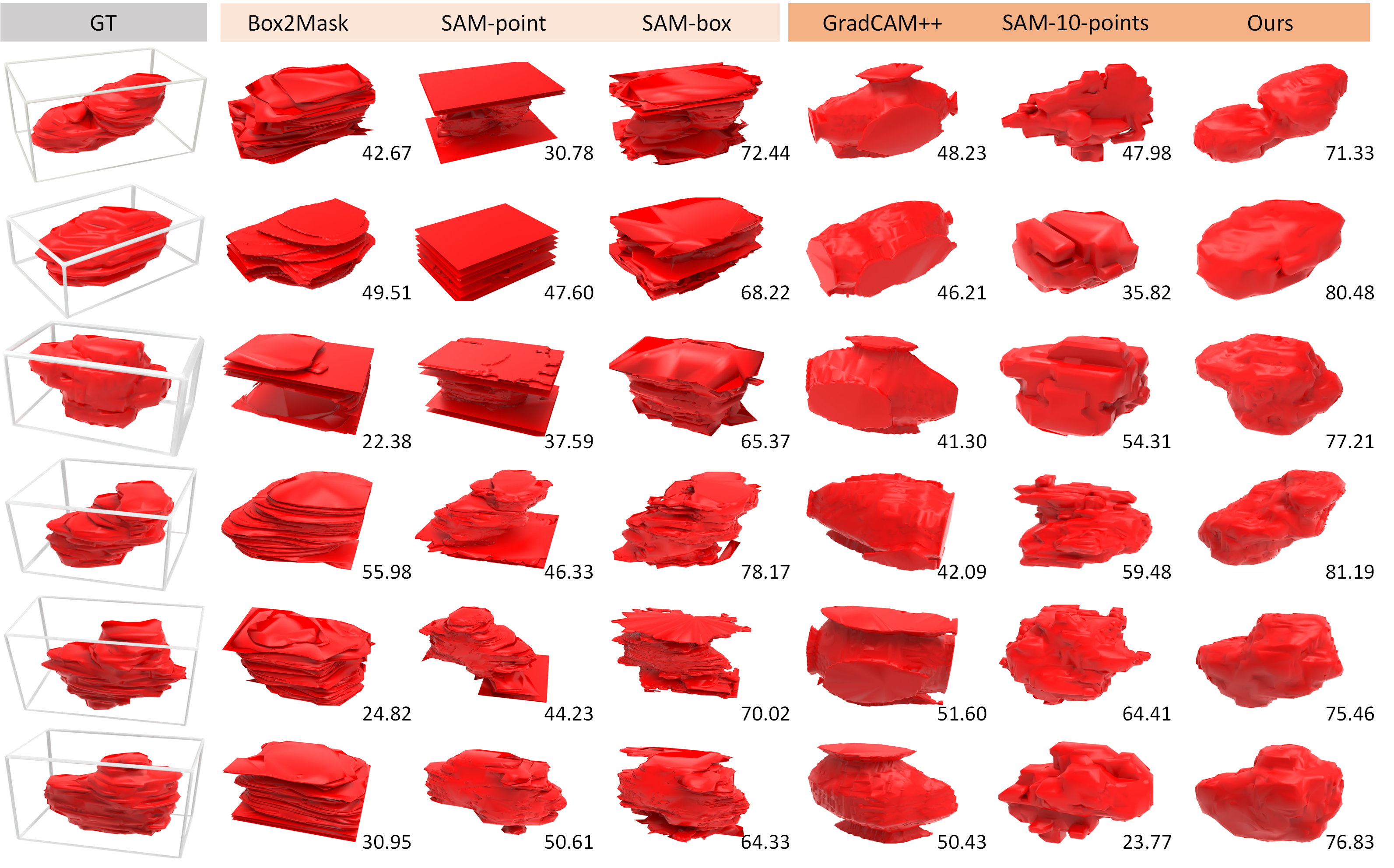}
	\caption{Visualization results of different methods (Columns 2-4: 2D-to-3D, Columns 5-7: 3D) on ABUS dataset. The right corner shows the DICE performance.}
	\label{fig:3d_vis}
\end{figure*}

\begin{figure*}[!t]
	\centering
	\includegraphics[width=1.0\linewidth]{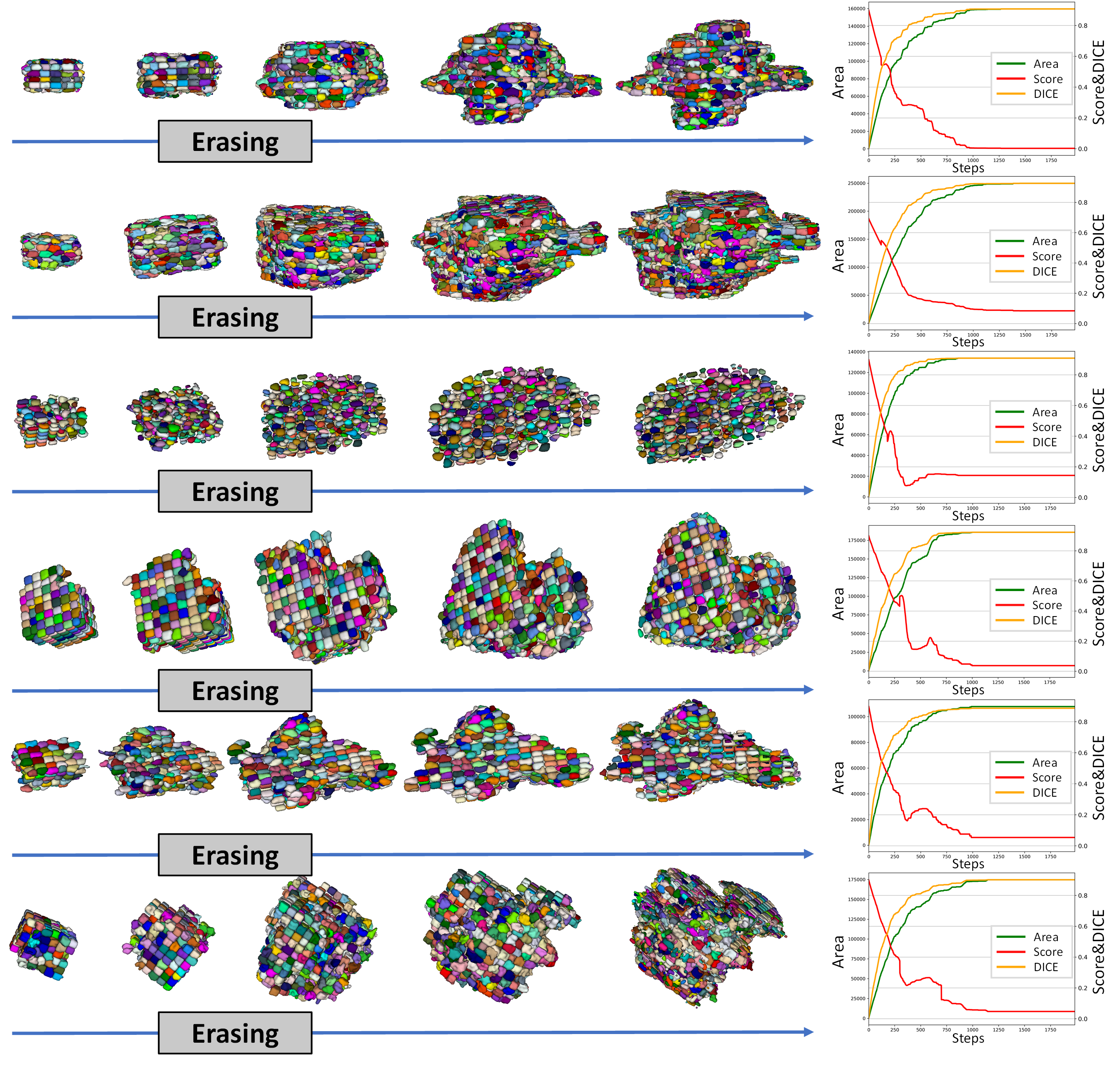}
	\caption{Six typical cases with good ABUS nodule segmentation results. For each case (row), columns 1-5 visualize the masks in different erasing steps. Column 6 shows the erasing curves including the DICE, the area, and the classification score at each step.}
	\label{fig:3D_result_step}
\end{figure*}

\begin{figure*}[!t]
	\centering
	\includegraphics[width=1.0\linewidth]{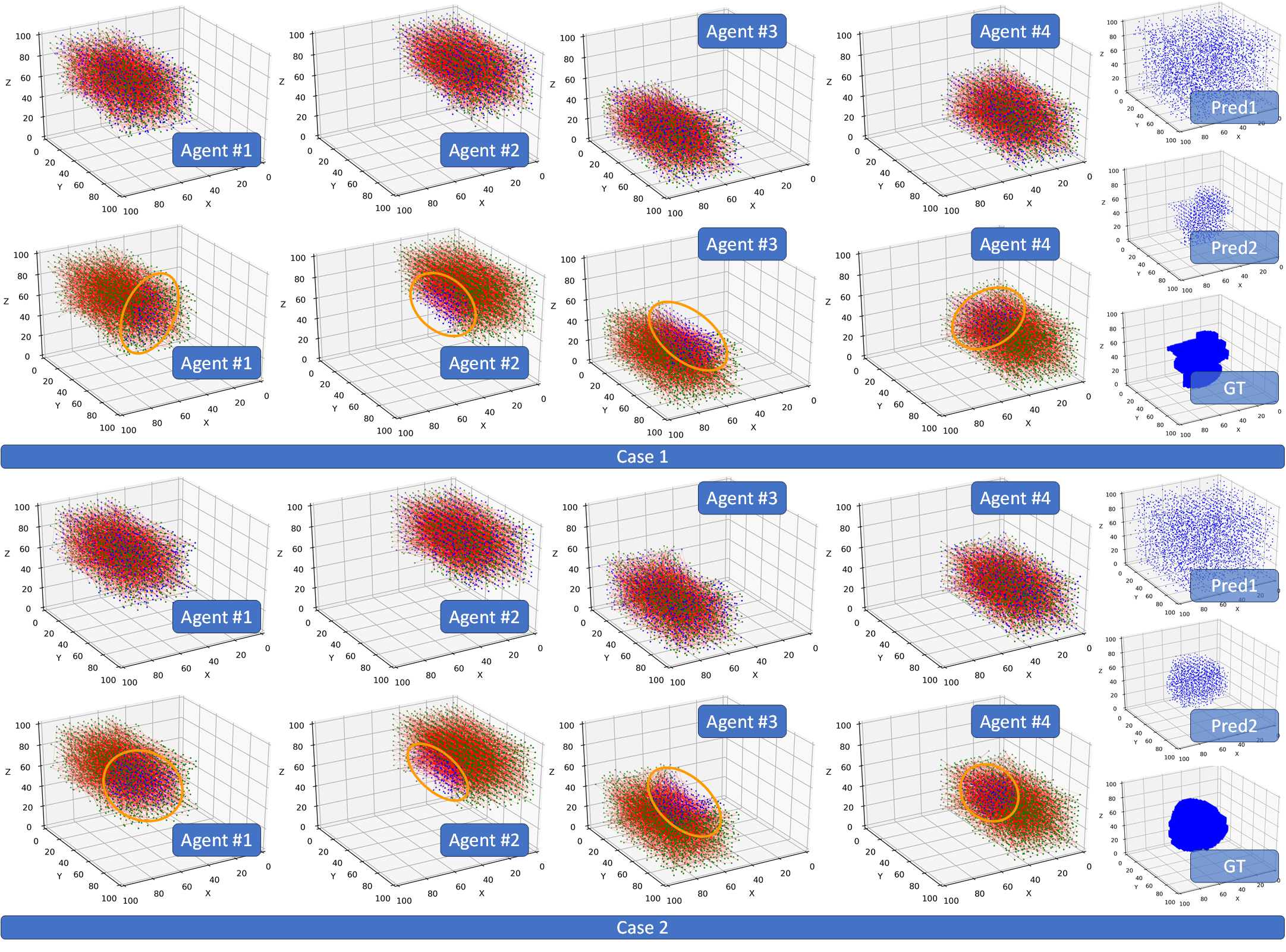}
	\caption{3D trajectory of multiple agents in two typical cases. The orange circles indicate that the erasing of the agents is concentrated and close to the center.}
	\label{fig:3D_traj}
\end{figure*}

\subsection{Supervised-based Method Comparison on BUS Dataset}
In WSS tasks, fully supervised methods often serve as an upper bound on performance since they typically have richer supervised information (e.g., mask vs. box).
In Table~\ref{tab:sup-bus}, we compared our method with six common and state-of-the-art approaches, including Unet, nnU-Net, Transunet, Swin-unet, U-mamba, and Swin-Umaba, on the BUS dataset.
We highlight that our work has vital clinical significance compared to supervised-based methods due to the following main reasons:

1) Our method achieves comparable global regional similarity (DICE) but significantly better boundary performance (HD) than the supervised-based methods.
The boundary of breast nodules is one of the key attributes to determine the grade of benign/malignant and BI-RADS.
Thus, boundary accuracy for breast nodule segmentation is very important in clinical reality.
As shown in Table~\ref{tab:sup-bus} (Testing-A), the DICE performance of various methods is very close, the difference is within 2\%.
Three of the six competitors obtained better average DICE performance than \textit{ours}.
Based on the paired t-test, only the improvement by nnU-Net and Swin-unet has statistical significance (p$<$0.05), and the results of Unet and \textit{ours} have no significant difference.
However, for HD metrics, due to the proposed superpixel-based \textit{Environment} for boundary prior extraction, \textit{ours} achieves significant improvement over all other methods (p$<$0.05).

(2) Our proposed approach shows better generalization ability on an external BUS dataset (Testing-B) than the supervised-based methods.
As shown in Table~\ref{tab:sup-bus} (Testing-B), \textit{Ours} outperforms competitors on all metrics with statistical significance (p$<$0.05).
This good generalization ensures the stable deployment of our method to complex clinical scenarios including different centers, devices, and imaging protocols.

(3) Under the same annotation cost (AC), our method achieves better performance than Unet and nnU-Net (Testing-A).
Note that for an experienced expert, annotating one nodule mask requires $\sim$2 minutes, while labeling the box needs about 10 seconds.
As shown in Table~\ref{tab:time_semi}, simply ignoring the consuming time for data loading, data switching, expert resting, etc, one hour can bring 30 masks and 360 boxes (AC=1), five hours can obtain 180 masks and 1800 boxes (AC=5), and annotating all masks requires about 80 hours (AC=80).
See lines \textit{Unet}, \textit{nnU-Net}, and \textit{Ours} (AC=1\&5), \textit{Ours} achieves significantly better performance than Unet and nnU-Net (p$<$0.05).
This illustrates the high efficiency of our approach, especially in clinical scenarios where annotation time and resources are limited and a segmentation algorithm needs to be developed quickly.

(4) Our method plays an efficient tool in assisting deep models via semi-supervised learning.
In Table~\ref{tab:time_semi}, we conducted several experiments to evaluate the power of our proposed model in boosting the supervised-based segmentation (i.e., Unet and nnU-net) in a semi-supervised manner.
Taking \textit{(Semi) Unet} and \textit{(Semi) nnU-net} (AC=1) for example, we first obtained the models trained on 30 images.
Since we have 2401 training cases, the trained models can generate pseudo masks for the remaining 2371 images.
We then combined the images with ground truths (30) and generated masks (2371) for semi-supervised training. 
The total epoch is 100, and we updated the pseudo labels every 10 epochs for iterative optimization.
For \textit{(Semi) Unet+Ours} and \textit{(Semi) nnU-net+Ours}, we used the model trained by our proposed method to generate higher-quality pseudo labels than \textit{Unet} and \textit{nnU-net} (DICE: 80.67\% vs. 65.18\%/70.44\%, p$<$0.05).
Other implementation details are consistent with the above settings for fair comparisons.
We found that under both \textit{AC=1} and \textit{AC=5}, using pseudo labels provided by our model show significantly better performance than by Unet and nnU-net, respectively (all evaluation metrics, p$<$0.05).
Moreover, \textit{(Semi) nnU-net+Ours} (AC=5) outperforms Unet and achieves comparable performance with nnU-Net (both require AC=80, $\sim$16$\times$ cost).
We also consider paired t-tests and find that p$<$ 0.05 for JAC and HD (\textit{(semi) nnU-Net+Ours} vs. \textit{nnU-Net}); however, for DICE and ASD metrics, the two methods have no statistical difference.
This further proves that our approach can enhance existing models via semi-supervised learning, while greatly reducing the cost of labeling.

\subsection{Foundation Model Comparison on BUS Dataset}
Since the latest development of foundation models, we also included different segment anything model (SAM) based methods prompted by \textit{box} for a fair comparison.
Specifically, except for the original SAM~\citep{kirillov2023segment, mazurowski2023segment} without specific medical fine-tuning, we supplemented MedSAM~\citep{ma2024u} and SAM-Med~\citep{huang2024segment} trained using medical images (including the BUS training set).
We also prepare different experimental settings, including \textit{Tight box} and \textit{0-10 pixels} box shifting, on both in-domain Testing-A and out-of-domain Testing-B datasets (refer to Table~\ref{tab:SAM-WSS}).
In the box-shifting scenario, we further report the results of RoBox-SAM~\citep{huang2024robust}, which is the latest solution for solving the box prompt shifting problem commonly in SAM.
Results in Table~\ref{tab:SAM-WSS} show that: (1) Under the tight box prompts, our proposed method outperforms the foundation models for both Testing-A and Testing-B datasets in most evaluated metrics and (2) Our approach copes well with box shiftings, which are common in real-world scenarios. However, foundation models will face performance degradation even with a slight level of 0-10 pixel shifts.

\subsection{Qualitative analysis on BUS dataset}
The visualization results are depicted in Figure~\ref{fig:result_2d}.
The results indicate that \textit{Ours} outperforms the strong competitors, including traditional methods, CAM- and box-based WSS approaches, and the SAM-based foundation model. 
Besides, \textit{Ours} can achieve comparable performance with the fully-supervised method (Unet) and even excels in certain instances (e.g., row 1 in Figure~\ref{fig:result_2d}).
Furthermore, we present the input images, erasing process, flipped images, predictions \& GTs, and erasing curves for five typical cases in Figure~\ref{fig:2D_result_step}.
The green arrows in the erasing process highlight the region erased by the agents at different steps.
Nodules in the input images (column 1) are erased, resulting in obtaining the flipped images (normal-like). 
Subsequently, the final erased region is extracted as the prediction, which closely resembles the GT (see column 6).
The curves depicted in the last part illustrate the relationship between the size of the erased area (green), the DICE metric (yellow), and the nodule classification score (red).
It is evident that as the erasing progresses, both the DICE and the erased area show a simultaneous increase, indicating a synchronous variation.
Moreover, the classification score exhibits a continuous decline, with the classification tag transitioning from ``nodule" to ``normal tissue".
This observation validates the effectiveness of our proposed \textit{Flip Learning} approach.

\subsection{Method Comparison on ABUS Dataset}
Due to the absence of labels for nodule masks in our ABUS training and validation sets, it is not feasible to conduct fully-supervised methods for comparison. Therefore, we present the supervised results from other studies on ABUS nodule segmentation in Table~\ref{tab:abus_comparison}. 
This table includes details on the dataset (such as the number of cases, volumes, and nodules) and the segmentation performance metrics (DICE and JAC). We consider this information to offer an approximate upper limit for the performance of ABUS nodule segmentation.
The evaluation values $a_b$ denote the mean ($a$) and standard deviation ($b$). An asterisk (*) indicates that the information has not been provided in the respective paper.
The most recent research by~\citet{pan2023gaussian} achieved the highest DICE among all the methods, 
Therefore, their performance can be considered as the theoretical upper limit of nodule segmentation in ABUS images.

Table~\ref{tab:abusweak} shows the results of different WSS-based methods and pre-trained foundation models (e.g., SAM, SAM2, etc.).
\textit{2D to 3D} represents that using 2D networks to handle the slices, and combine the 2D results to form the 3D one.
\textit{Video} indicates the SAM2-based methods to model the volumes as videos.
\textit{3D} means that directly using 3D networks to output volumetric results. 
Among the \textit{2D to 3D} based methods, the CAM-based methods achieve a poor mean DICE with a range of 23.49\% to 27.59\%. 
Box-based approaches can improve the results, and box-driven SAM performs best among competitors (DICE: 65.36\%).
However, they all handle 2D slices individually, limiting their practicability to 3D data.

Recently, SAM 2 has been introduced and its video segmentation capabilities have been leveraged to segment volumes by first converting them into videos~\citep{ma2024segment2}.
Different from SAM2 concentrated on natural images, MedSAM2 and SAM2-MedIV are finetuned using medical ultrasound images via different strategies (i.e., memory enhancement and transfer learning).
Specifically, given a 2D box, '-F' represents that we start at the first slice with the box prompt, and infer till the last frame (forward).
'-M' means that the middle frame is selected as the starting point, with bi-direction inference, i.e., middle-to-first (reverse) and middle-to-end (forward).
SAM2-MedIV-M shows better DICE\&JAC performance than SAM-box, validating the superiority of video-based methods.
However, they still perform worse than our method, with a DICE gap of about 7\%.

We also implemented the 3D-based approaches, which can capture spatial information well and may achieve better performance than the corresponding 2D-to-3D methods.
For example, 3D GradCAM++ outperforms 2D GradCAM++ by 22.10\% in DICE.
In addition, 3D-SAM with only one point prompt performs better than 2D-SAM with slice-wise point prompts, especially in the distance-related metrics (i.e., HD and ASD).
As the number of points increases, the segmentation performance of 3D SAM will improve, but it will gradually stabilize at about 10 point prompts (DICE: 48.03\%).
Thus, their results still have much room for progress compared to the fully-supervised methods.
In the last row, the results of \textit{ours} outperform all competitors, highlighting the effectiveness of the \textit{Flip learning} mechanism in 3D images.

Qualitative results are shown in Figure~\ref{fig:3d_vis}.
All other approaches (columns 2-6) suffer from poor inter-frame continuity or unreasonable shape.
In contrast, \textit{Ours} shows better boundary and overall shape segmentation.
3D supervoxel-level prediction in different erasing steps and erasing curves can be found in Figure~\ref{fig:3D_result_step}.
See the DICE curves (yellow), the final predicted masks can achieve high overlap with GTs (DICE$>$0.9).
We also present the 3D moving trajectory of agents in Figure~\ref{fig:3D_traj}. 
In each case, the first row shows the results obtained from the RL \textit{w/o} training, and the second row shows the results of the best-performed model (\textit{ours}). 
For agents $\#$1-$\#$4, the red lines visualize the moving paths of the agents. 
Here, we use the center points of supervoxels to represent the agents' locations. 
The green and blue points show the action \textit{passing} and \textit{erasing}, respectively. 
\textit{Pred1} (\textit{w/o} training) and \textit{Pred2} (\textit{w} training) in the most right part show all points with \textit{erasing} actions.
These point clouds highlight that the post-training agents can take correct actions and obtain segmentation predictions close to the \textit{GTs}.

We also validate the performance robustness of different methods (2D-to-3D: \textit{BoxMask}, 3D: \textit{GradCAM++, SAM-10-points, and Ours}) across ABUS nodules with different BI-RADS levels.
In Table~\ref{tab:biradsrobustness}, we provide the average DICE of different types of nodules (BI-RADS2-4), and also the robustness calculated by standard deviations ($R_{std}$). 
A smaller $R_{std}$ signifies better stability of the method.
It can be observed that as the BI-RADS grades of the nodules increase, the segmentation performance of different methods generally decreases.
\textit{Ours} achieves better mean DICE results across all three types of nodules ($Avg$: 75.48\%) while maintaining optimal stability ($R_{std}$: 0.67).

\begin{table}[!h]
  \centering
  \caption{Performance robustness across different BI-RADS levels. The best results are shown in bold.}
  \resizebox{0.48\textwidth}{!}{
    \begin{tabular}{cccccc}
    \toprule
    \multirow{2}[4]{*}{} & \multicolumn{3}{c}{DICE$\uparrow$} & \multirow{2}[4]{*}{Avg$\uparrow$} & \multirow{2}[4]{*}{R$_{std}$$\downarrow$} \\
\cmidrule{2-4}          & BI-RADS2 & BI-RADS3 & BI-RADS4 &       &  \\
    \midrule
    2D-to-3D BoxMask & 56.09  & 54.17  & 50.26  & 52.77  & 2.43  \\
    3D GradCAM++ & 52.24  & 49.81  & 43.92  & 47.63  & 3.49  \\
    3D SAM-10-points & 49.86  & 48.19  & 47.22  & 48.03  & 1.09  \\
        Ours  & \textbf{76.62}  & \textbf{75.53}  & \textbf{75.02}  & \textbf{75.48}  & \textbf{0.67}  \\
    \bottomrule
    \end{tabular}}%
  \label{tab:biradsrobustness}%
\end{table}%

\subsection{Ablation Studies on the RL Strategies}
\label{sec:ablation}
Tables~\ref{tab:bus_ablation} and~\ref{tab:abus_ablation} test the contribution of each proposed strategy.
\textit{Baseline} represents our previous MICCAI work~\citep{huang2021flip}.
It is a MARL framework with non-regularized \textit{environment}, index-order-based \textit{action}, CSR and IDR1 \textit{rewards}, and coarse-to-fine (C2F) \textit{learning strategy}.
\textit{Baseline-E} means the environment was regularized to a fixed size. 
\textit{Baseline-A} reports the results of the optimized path traverse of agents.
The agents will traverse the environment following the new index map, generated based on the distance from each superpixel/supervoxel to the center one in an inter-to-outer principle.
\textit{Baseline-R} shows the results of adding IDR2.
\textit{Baseline-MCL} and \textit{Baseline-PCL} are two different one-stage learning strategies: mixed curriculum learning and progressive curriculum learning.
Specifically, \textit{-MCL} means that before each data input to the network, we randomly choose the curriculum, i.e., the number of superpixels/supervoxels, to encode the environment.
Thus, the agents learn in the environment containing mixed curricula in an unstructured manner.
\textit{-PCL} means that the agents' learning follows an easy-to-hard manner, similar to progressive pre-training for reducing the difficulty of model learning. 

Experiments validate that regularizing the environment can improve the segmentation performance (\textit{Baseline vs. Baseline+E}).
The size regularization strategy avoids the agents interacting with environments of large scale changes, thus making the learning process more stable and efficient.
See \textit{Baseline} and \textit{Baseline-A}, the inter-to-outer traversal rule can improve the model accuracy in both datasets.
It proves that, instead of interacting without any attention and moving order tendency, the proposed traversal method can effectively help agents first focus on learning the potential target area using prior distance knowledge.
Adding IDR2 also optimizes overall performance, since this reward signal drives agents concerned about the differences between the potential foreground and background.
It can be observed that both \textit{-MCL} and \textit{-PCL} achieve better segmentation performance than \textit{-C2F} for BUS and ABUS datasets.
Especially, \textit{-PCL} outperforms the other two strategies with 1.52\%/1.33\% and 2.39\%/2.14\% in the DICE metric for the BUS and ABUS datasets, respectively.
This further demonstrates the effectiveness of the PCL strategy.
The last rows (\textit{Ours}) of Tables~\ref{tab:bus_ablation} and~\ref{tab:abus_ablation} shows the results of equipping the \textit{Baseline} with all the strategies, including \textit{-E}, \textit{-A}, \textit{-R} and \textit{-PCL}.
It shows that leveraging all the strategies together will further improve all the evaluation metrics.
The shown significant improvement in Tables can better prove the contributions of our proposed strategies, especially for \textit{-E}, \textit{-A}, \textit{-R} and \textit{-PCL}.

Besides, we tried different common superpixel/supervoxel generation algorithms for the environment, including LSC~\citep{li2015superpixel} and SEEDS~\citep{van2015seeds}. We replaced the SLIC algorithm with LSC and SEEDS within both \textit{Baseline} and \textit{Ours}, and the results are shown in Table~\ref{tab:superpv}.
It can be observed that both \textit{Baseline}$\&$\textit{Ours} equipped with SLIC obtain better average performance than the competitors on all metrics, but most improvements are not significant (p$>$0.05).
This illustrates that compared to LSC and SEEDS, SLIC is slightly more suitable for our tasks of BUS/ABUS nodule boundary extraction.
Results also highlight that our method is general to achieve competitive and comparable segmentation performance under different superpixel/supervoxel generation algorithms.

Moreover, we comprehensively analyze the reward functions in Table~\ref{tab:reward-ablation}.
It can be observed that in both BUS and ABUS datasets, compared to integrating CSR only, all evaluation metrics drop significantly without CSR.
The satisfactory results with CSR further ensure its fundamental and important role in driving the erasing process and achieving good segmentation performance.
We also notice that the proposed rewards (IDR1 and IDR2) can model distribution relationships well and improve generalization ability. 
We use \textit{$\Delta$DICE} to describe the average DICE performance drop from \textit{Testing-A} to \textit{Testing-TB}.
As shown in Table~\ref{tab:reward-ablation}, without IDR1 and IDR2, our generalization performance is unsatisfactory ($\Delta$DICE=3.01).
Otherwise, adding IDR1, IDR2 or their combination can enhance the model generalization ability.
Besides, IDR2 often contributes more to the improved generalization than IDR1.
This further validates the importance of IDR2 for its explicit relative distribution learning in improving generalization ability.

\begin{table}[!t]
	\centering
        \scriptsize
	\caption{Ablation studies on BUS dataset (\textit{Testing-A}). The best results are shown in bold. $^*$, $^{**}$, $^{***}$ mean the p-value of paired t-tests between ablation methods and baseline less than 0.05, 0.01, and 0.001, respectively. No $^*$ marks represent no significant difference.}
    \resizebox{0.48\textwidth}{!}{
	\begin{tabular}{ccccc}
		\toprule
		& DICE$\uparrow$  & JAC$\uparrow$   & HD$\downarrow$    & ASD$\downarrow$ \\
		\midrule
    Baseline & 89.71±7.21 & 84.99±9.52 & 16.11±15.87 & 6.71±6.22 \\
    Baseline+E & 91.44±4.89$^{**}$ & 85.83±9.88$^{**}$ & 14.67±14.98$^{***}$ & 4.05±4.01$^{***}$ \\
    Baseline+A & 91.07±6.97$^{**}$ & 86.05±7.54$^{***}$ & 15.28±15.43$^{**}$ & 4.73±5.13$^{***}$ \\
    Baseline+R & 90.87±7.04$^{*}$ & 85.57±8.14$^{**}$ & 15.44±15.28$^{**}$ & 4.56±4.71$^{***}$ \\
    Baseline+MCL & 89.90±8.26* & 84.71±8.77 & 17.14±13.57 & 5.18±3.58$^{**}$ \\
    Baseline+PCL & 91.23±6.92$^{**}$ & 85.91±7.09$^{***}$ & 15.24±14.91$^{**}$ & 3.74±2.98$^{***}$ \\
		\midrule
		Ours  & \textbf{92.37±4.22$^{***}$} & \textbf{86.09±6.80$^{***}$} & \textbf{14.19±13.89$^{***}$} & \textbf{3.69±2.50$^{***}$} \\
		\bottomrule
	\end{tabular}}%
	\label{tab:bus_ablation}%
\end{table}%

\begin{table}[!t]
	\centering
        \scriptsize
	\caption{Ablation studies on ABUS dataset. The best results are shown in bold. $^*$, $^{**}$, $^{***}$ mean the p-value of paired t-tests between ablation methods and baseline less than 0.05, 0.01, and 0.001, respectively.}
    \resizebox{0.48\textwidth}{!}{
	\begin{tabular}{ccccc}
		\toprule
		& DICE$\uparrow$  & JAC$\uparrow$   & HD$\downarrow$  & ASD$\downarrow$   \\
		\midrule
    Baseline & 69.32±16.21 & 52.67±18.92 & 28.67±35.22  & 5.87±4.18 \\
    Baseline+E & 71.42±16.04$^{**}$ & 54.78±15.52$^{***}$  & 21.72±32.18$^{***}$  & 4.76±3.21$^{**}$ \\
    Baseline+A & 73.27±17.12$^{***}$  & 55.58±14.17$^{***}$ & 29.15±24.67$^{***}$  & 4.56±3.08$^{***}$ \\
    Baseline+R & 72.87±14.11$^{***}$  & 55.99±15.01$^{***}$  & 28.86±25.71$^{***}$  & 4.03±3.17$^{***}$ \\
    Baseline+MCL & 69.57±15.23$^{*}$ & 54.52±13.84$^{**}$  & 26.99±23.12$^{***}$  & 5.23±3.25$^{**}$ \\
    Baseline+PCL & 71.71±14.77$^{***}$  & 56.71±14.72$^{***}$  & 23.52±22.86$^{***}$  & 4.27±2.01$^{***}$ \\
    \midrule
    Ours  & \textbf{75.48±12.67$^{***}$}  & \textbf{61.22±10.41$^{***}$}  & \textbf{20.08±18.16$^{***}$}  & \textbf{3.09±1.83$^{***}$} \\
		\bottomrule
	\end{tabular}}%
	\label{tab:abus_ablation}%
\end{table}%

 \begin{table}[!t]
  \centering
  \small
  \caption{Model performance on different superpixel or supervoxel generation algorithms.}
  \resizebox{0.48\textwidth}{!}{
    \begin{tabular}{ccccccc}
    \toprule
          &       &       & DICE$\uparrow$  & JAC$\uparrow$   & HD$\downarrow$    & ASD$\downarrow$ \\
    \midrule
    \multicolumn{1}{c}{\multirow{6}[4]{*}{2D (BUS)}} & \multirow{3}[2]{*}{Baseline} & LSC   & 89.61±7.53 & 84.83±9.93 & 16.78±16.07 & 7.22±6.05 \\
          &       & SEEDS & 89.63±7.49 & 84.74±9.78 & 16.43±16.21 & 7.05±6.14 \\
          &       & SLIC  & 89.71±7.21 & 84.99±9.52 & 16.11±15.87 & 6.71±6.22 \\
\cmidrule{2-7}          & \multirow{3}[2]{*}{Ours} & LSC   & 92.29±4.17 & 85.80±6.92 & 14.77±13.76 & 4.20±2.79 \\
          &       & SEEDS & 91.89±5.66 & 85.74±7.14 & 15.63±14.21 & 4.17±3.02 \\
          &       & SLIC  & 92.37±4.22 & 86.09±6.80 & 14.19±13.89 & 3.69±2.50 \\
    \midrule
          &       &       & DICE$\uparrow$  & JAC$\uparrow$   & HD$\downarrow$    & ASD$\downarrow$ \\
    \midrule
    \multicolumn{1}{c}{\multirow{6}[4]{*}{3D (ABUS)}} & \multirow{3}[2]{*}{Baseline} & LSC   & 69.15±16.70 & 51.93±18.98 & 29.05±34.71 & 6.05±4.07 \\
          &       & SEEDS & 69.30±17.11 & 52.47±19.02 & 28.84±34.55 & 6.17±4.20 \\
          &       & SLIC  & 69.32±16.21 & 52.67±18.92 & 28.67±35.22 & 5.87±4.18 \\
\cmidrule{2-7}          & \multirow{3}[2]{*}{Ours} & LSC   & 75.45±12.66 & 61.10±10.19 & 21.23±18.08 & 3.21±1.99 \\
          &       & SEEDS & 75.40±12.55 & 61.18±10.25 & 21.11±17.98 & 3.15±2.01 \\
          &       & SLIC  & 75.48±12.67 & 61.22±10.41 & 20.08±18.16 & 3.09±1.83 \\
    \bottomrule
    \end{tabular}}%
  \label{tab:superpv}%
\end{table}%

\begin{table}[!t]
  \centering
  \small
  \caption{Reward analysis on BUS and ABUS datasets.}
  \resizebox{0.48\textwidth}{!}{
    \begin{tabular}{ccccccc}
    \toprule
    \multicolumn{3}{c}{\multirow{2}[4]{*}{Reward}} & \multicolumn{4}{c}{BUS (Testing-A)} \\
\cmidrule{4-7}    \multicolumn{3}{c}{}  & \multirow{2}[4]{*}{DICE$\uparrow$} & \multirow{2}[4]{*}{JAC$\uparrow$} & \multirow{2}[4]{*}{HD$\downarrow$} & \multirow{2}[4]{*}{ASD$\downarrow$} \\
\cmidrule{1-3}    CSR   & IDR1  & IDR2  &       &       &       &  \\
    \midrule
    \ding{51}     & \ding{55}     & \ding{55}     & 91.15±7.17 & 85.99±9.47 & 16.07±18.82 & 6.55±5.79 \\
    \ding{55}     & \ding{51}     & \ding{55}     & 70.60±13.24 & 61.89±12.92 & 21.78±20.95 & 8.63±7.28 \\
    \ding{55}     & \ding{55}     & \ding{51}     & 69.92±14.86 & 60.33±13.19 & 22.02±21.38 & 8.51±7.16 \\
    \ding{55}     & \ding{51}     & \ding{51}     & 71.32±13.72 & 62.44±12.56 & 20.11±19.54 & 8.14±6.94 \\
    \ding{51}     & \ding{51}     & \ding{55}     & 91.62±6.89 & 85.74±9.18 & 15.87±18.32 & 4.78±5.26 \\
    \ding{51}     & \ding{55}     & \ding{51}     & 91.71±6.14 & 85.95±7.88 & 15.01±16.87 & 5.09±4.78 \\
    \ding{51}     & \ding{51}     & \ding{51}     & 92.37±4.22 & 86.09±6.80 & 14.19±13.89 & 3.69±2.50 \\
    \midrule
    \multicolumn{3}{c}{\multirow{2}[4]{*}{Reward}} & \multicolumn{4}{c}{BUS (Testing-B)} \\
\cmidrule{4-7}    \multicolumn{3}{c}{}  & \multirow{2}[4]{*}{DICE$\uparrow$} & \multirow{2}[4]{*}{JAC$\uparrow$} & \multirow{2}[4]{*}{HD$\downarrow$} & \multirow{2}[4]{*}{ASD$\downarrow$} \\
\cmidrule{1-3}    CSR   & IDR1  & IDR2  &       &       &       &  \\
    \midrule
    \ding{51}     & \ding{55}     & \ding{55}     & 88.14±8.92 & 82.68±10.83 & 17.84±19.05 & 7.23±6.11 \\
    \ding{55}     & \ding{51}     & \ding{55}     & 68.84±13.76 & 59.76±13.52 & 23.02±21.77 & 9.05±7.95 \\
    \ding{55}     & \ding{55}     & \ding{51}     & 68.93±14.99 & 59.54±14.12 & 23.18±22.05 & 8.99±8.33 \\
    \ding{55}     & \ding{51}     & \ding{51}     & 70.53±14.16 & 60.99±13.01 & 21.14±20.20 & 8.53±7.08 \\
    \ding{51}     & \ding{51}     & \ding{55}     & 91.07±7.85 & 84.83±11.14 & 17.54±19.11 & 5.09±5.35 \\
    \ding{51}     & \ding{55}     & \ding{51}     & 91.34±5.97 & 85.19±9.74 & 16.61±17.85 & 5.23±5.10 \\
    \ding{51}     & \ding{51}     & \ding{51}     & 92.15±5.61 & 85.66±9.36 & 16.30±17.40 & 4.07±4.24 \\
    \midrule

    \multicolumn{3}{c}{\multirow{2}[4]{*}{Reward}} & \multicolumn{4}{c}{ABUS} \\
\cmidrule{4-7}    \multicolumn{3}{c}{}  & \multirow{2}[4]{*}{DICE$\uparrow$} & \multirow{2}[4]{*}{JAC$\uparrow$} & \multirow{2}[4]{*}{HD$\downarrow$} & \multirow{2}[4]{*}{ASD$\downarrow$} \\
\cmidrule{1-3}    CSR   & IDR1  & IDR2  &       &       &       &  \\
    \midrule
    \ding{51}     & \ding{55}     & \ding{55}     & 73.83±13.88 & 57.86±14.92 & 22.71±21.99 & 4.98±2.37 \\
    \ding{55}     & \ding{51}     & \ding{55}     & 58.99±19.41 & 42.14±19.53 & 36.21±25.78 & 8.01±5.63 \\
    \ding{55}     & \ding{55}     & \ding{51}     & 57.68±18.50 & 40.89±20.13 & 35.01±24.80 & 8.13±6.05 \\
    \ding{55}     & \ding{51}     & \ding{51}     & 62.75±17.76 & 47.81±18.02 & 32.26±23.77 & 7.14±6.28 \\
    \ding{51}     & \ding{51}     & \ding{55}     & 74.17±15.62 & 58.96±12.88 & 23.91±20.84 & 4.29±2.45 \\
    \ding{51}     & \ding{55}     & \ding{51}     & 74.86±13.24 & 60.73±11.82 & 22.14±19.56 & 3.88±2.30 \\
    \ding{51}     & \ding{51}     & \ding{51}     & 75.48±12.67 & 61.22±10.41 & 20.08±18.16 & 3.09±1.83 \\
    \bottomrule
    \end{tabular}}%
  \label{tab:reward-ablation}%
\end{table}%

\begin{table*}[!t]
	\centering
        \footnotesize
	\caption{Ablation studies on the number of agents. The best results are shown in bold.}
 % \resizebox{0.6\textwidth}{!}{
        \setlength{\tabcolsep}{4mm}{
	\begin{tabular}{ccccccc}
		\toprule
		& \multicolumn{2}{c}{BUS (Testing-A)} & \multicolumn{2}{c}{BUS (Testing-B)} & \multicolumn{2}{c}{ABUS} \\
		\midrule
		Numbers of Agents & DICE$\uparrow$  & HD$\downarrow$    & DICE$\uparrow$  & HD$\downarrow$    & DICE$\uparrow$  & HD$\downarrow$ \\
		\midrule
		\#1   & 88.12±5.87 & 19.28±14.19 & 87.78±6.17 & 21.05±18.66 & 67.57±11.55 & 27.19±23.61 \\
		\#2   & \textbf{92.37±4.22} & \textbf{14.19±13.89} & \textbf{92.15±5.61} & \textbf{16.30±17.40} & 72.45±11.78 & 24.24±18.99 \\
		\#3   & 91.84±5.05 & 14.41±15.02 & 91.08±6.01 & 18.43±16.23 & 73.41±12.72 & 22.36±20.48 \\
		\#4   & 90.58±4.33 & 15.88±14.78 & 90.43±6.24 & 19.08±17.11 & \textbf{75.48±12.67} & \textbf{20.08±18.16} \\
		\#5   & 90.27±4.59 & 16.94±14.78 & 89.92±5.87 & 20.23±18.91 & 74.05±11.92 & 21.68±20.72 \\
		\#6   & 89.68±5.72 & 18.16±16.71 & 89.74±6.87 & 20.48±18.01 & 73.17±13.05 & 21.58±21.98 \\
		\bottomrule
	\end{tabular}}%
	\label{tab:agent_number}%
\end{table*}%

\begin{table*}[!htbp]
	\centering
        \footnotesize
	\caption{Ablation studies on different curriculum settings. The best results are shown in bold.}
 % \resizebox{0.9\textwidth}{!}{
        \setlength{\tabcolsep}{4mm}{
	\begin{tabular}{cccccccc}
		\toprule
		&       & \multicolumn{2}{c}{BUS (Testing-A)} & \multicolumn{2}{c}{BUS (Testing-B)} & \multicolumn{2}{c}{ABUS} \\
		\midrule
		Curriculum     & Range & DICE$\uparrow$  & HD$\downarrow$    & DICE$\uparrow$  & HD$\downarrow$    & DICE$\uparrow$  & HD$\downarrow$ \\
		\midrule
		$c_1$-$c_2$     & 100-1000 & 86.41±8.71 & 19.36±15.89 & 85.79±9.12 & 21.41±20.58 & 63.23±21.29 & 28.22±23.17 \\
		$c_1$-$c_3$     & 100-2000 & \textbf{92.37±4.22} & \textbf{14.19±13.89} & \textbf{92.15±5.61} & \textbf{16.30±17.40} & 65.31±19.74 & 26.35±21.78 \\
		$c_1$-$c_4$     & 100-5000 & 89.02±7.18 & 16.88±14.92 & 88.68±9.22 & 19.76±18.97 & 67.11±15.92 & 23.99±22.08 \\
		$c_1$-$c_5$    & 100-10000 & 85.09±8.20 & 20.17±16.23 & 84.39±8.72 & 22.94±19.11 & \textbf{75.48±12.67} & \textbf{20.08±18.16} \\
		$c_1$-$c_6$    & 100-15000 & 81.17±7.33 & 24.55±16.87 & 80.11±9.05 & 26.45±21.34 & 73.67±12.58 & 21.42±20.70 \\
		\bottomrule
	\end{tabular}}%
	\label{tab:CL_t_setting}%
\end{table*}%

\begin{table}[!t]
  \centering
  \scriptsize
  \caption{Comparison of model performance under different box shifting levels.}
  \resizebox{0.48\textwidth}{!}{
    \begin{tabular}{ccccccc}
    \toprule
& \multicolumn{2}{c}{BUS (Testing-A)} & \multicolumn{2}{c}{BUS (Testing-B)} & \multicolumn{2}{c}{ABUS} \\
    \midrule
          & DICE$\uparrow$  & HD$\downarrow$    & DICE$\uparrow$  & HD$\downarrow$    & DICE$\uparrow$  & HD$\downarrow$ \\
    \midrule
    \multicolumn{1}{c}{Tight box} & 92.37 & 14.19 & 92.15 & 16.31 & 75.48 & 20.08 \\
    0-10 pixels/voxels & 91.08 & 15.21 & 90.87 & 18.05 & 73.53 & 22.21 \\
    10-20 pixels/voxels & 89.55 & 17.43 & 89.17 & 18.99 & 72.17 & 24.39 \\
    20-30 pixels/voxels & 87.93 & 20.36 & 87.03 & 21.85 & 69.21 & 27.83 \\
    \bottomrule
    \end{tabular}}%
  \label{tab:robust}%
\end{table}%

\subsection{Robustness Evaluation for Box Shifts at Various Levels}
In the previous experiments, we tested our framework using the tight boxes.
Specifically, the nodules are tightly contained in the minimum bounding boxes.
However, it is impractical to give precise boxes for the nodules in clinical reality.
To assess the sensitivity of our methodology to the box, we conducted validation on various levels of box shifting, including 1) 0-10, 2) 10-20 and 3) 20-30 pixels/voxels.
This can better simulate human operations in reality.
As shown in Table~\ref{tab:robust}, we randomly shift the boxes \textit{three} times and report the average DICE and HD results.
The results indicate that our method maintains good performance even with 20-30 pixel or voxel shifts, achieving DICE of 87.93\%, 87.03\%, and 69.21\% for BUS (Testing-A), BUS (Testing-B), and ABUS, respectively.

\subsection{Quantitative Analysis of the Number of Agents}
The number of agents will influence the deep model in terms of learning ability, speed, and also the performance of final segmentation.
In a multi-agent setting, each agent interacts with a different and non-overlapping sub-environment.
Multiple agents will learn and share the general environment information, and extract the individual decision knowledge via the joint features.
Increasing the number of agents will speed up their interaction process, thus, saving both training and testing time.
However, excessive agents will lead to huge differences in the knowledge they learn.
Such inconsistent shared information may affect the learning stability of the model, ultimately resulting in a serious performance drop.

In Table~\ref{tab:agent_number}, we report the DICE and HD metrics obtained by RL models with 1-6 agent(s) in three testing sets, including BUS (Testing-A and Testing-B) and ABUS.
For a fair comparison, all experiments are equipped with the same strategies, i.e., \textit{-E, -A, -R} and \textit{-PCL}.
The only difference among these compared experiments is the number of agents.
Note that \textit{\#1} is essentially a SARL method, i.e., only one agent independently explores the environment without any knowledge sharing.
\textit{\#2}-\textit{\#6} are based on MARL, in which $\mathcal{K}$ agents share information about the environment with each other.
It can be seen in Table~\ref{tab:agent_number} that the SARL method (\textit{\#1}) shows the worst performance in all three experimental groups.
We conclude that only one agent with limited learning ability may difficult to explore efficiently in a huge environment.
Without shared information and guidance from other agents, the independent agent will easily get lost in the interaction with the environment.  
It can also be observed that adding one more agent (\textit{\#2}) can significantly help the model's learning, improving the segmentation performance.
Specifically, \textit{\#2} can boost 4.25\%, 4.37\% and 4.88\% on DICE metrics compared to \textit{\#1} in three testing sets, respectively.
The optimal number of agents is different for the BUS and ABUS datasets. 
For the BUS datasets, \textbf{two} agents working together achieve the best results among all settings.
For the ABUS testing set, the optimal number is \textbf{four}.
We suspect that learning in 3D space should be more challenging than in 2D, thus, more agents are required.
Furthermore, experiments have validated that there is no positive correlation between agent numbers and final performance.
See \textit{\#2} \textit{vs.} \textit{\#3,4,5,6} of BUS and \textit{\#4} \textit{vs.} \textit{\#5,6} of ABUS, learning with more agents will decrease the DICE and HD scores.
This may be caused by inconsistent shared knowledge and unstable learning of multiple agents.

\subsection{Quantitative Analysis of the PCL Settings}
PCL plays an important role in guiding agents to learn in the correct direction.
A suitable curriculum setting can encode the environment finely while ensuring it is not too difficult for the learning of agents.   
Specifically, the course difficulty (i.e., the number of superpixels/supervoxels) in our ablation study includes 100, 1000, 2000, 5000, 10000, and 15000.
We performed experiments with different curriculum settings in Table~\ref{tab:CL_t_setting}.
It shows that for BUS data, three curricula ($c_1$-$c_3$) with the hardest course having 2000 superpixels outperform the other settings in all the evaluation metrics.
Besides, for the ABUS dataset, results under the setting of $c_1$-$c_5$ reveal the best performance.
It is also noted that results among different curriculum settings have significant diversity, e.g., BUS (Testing-A): 81.17\%-92.37\% in DICE.
Thus, setting the appropriate parameter in PCL is vital for achieving satisfactory performance.

\section{Discussion and Conclusion}
In this research, we aim to conceptualize the task of nodule segmentation as an erasing task based on RL. Our proposed \textit{Flip Learning} framework, which operates under weak supervision, necessitates just a single box for nodule segmentation and is applicable and general to both 2D and 3D breast US images. 

Within the proposed framework, multiple agents are engaged in interacting with the regularized environment (i.e., target box) to erase the nodule and fill the area for tag flipping in classification.
The environment is represented by superpixels/supervoxels for preliminary boundary extraction and enhancement of learning efficiency. 
We meticulously devise one reward to steer the segmentation process and two rewards to prevent excessive segmentation.
Additionally, we introduce a PCL approach to assist the agents in transitioning from simpler to more challenging tasks.
Extensive experiments conducted on our extensive in-house BUS and ABUS datasets confirm the efficacy and resilience of our proposed \textit{Flip Learning} framework.

Our proposed framework holds promise for expansion into a generative framework. Following the erasure process, the modified image labeled as ``normal tissue" can be generated. Consequently, our model can create paired normal images by erasing nodules from abnormal images, thereby facilitating the creation of large datasets comprising paired healthy-anomalous images within the medical image analysis community. Ultimately, this will expedite the advancement of pertinent deep learning models, benefiting clinical diagnosis and treatment planning.

We plan to assess the generative capability of the proposed \textit{Flip Learning} and its potential applications in our subsequent study. In future work, we will explore the possibility of training an object detector to automatically provide nodule bounding boxes during testing. Furthermore, we will introduce innovative zero-shot segmentation techniques~\citep{huang2024foundation}, such as the medical SAM, to offer high-quality initial segmentation predictions. Such precise initialization can offer crucial cues to guide the agents in their erasing process and enhance efficiency. Subsequently, we may investigate various interaction approaches between agents and the environment, encompassing diverse action spaces, and flexible termination strategies, among others. Finally, we aim to extend the methodology to diverse segmentation tasks involving various organs/lesions or imaging modalities.

\section{Acknowledgment}
This work was supported by the National Natural Science Foundation of China (Nos. 12326619, 62171290, 62471305, 62101342); Science and Technology Planning Project of Guangdong Province (No. 2023A0505020002); Frontier Technology Development Program of Jiangsu Province (No. BF2024078), Construction Fund of Medical Key Disciplines of Hangzhou (No. oo20200457); Hangzhou Biomedical and Health Industry Development Support Special Project (No. 2021WJCY091); Guangdong Basic and Applied Basic Research Foundation (No. 2023A1515012960); Royal Academy of Engineering INSILEX Chair under Grant CiET1919/19; UK Research and Innovation (UKRI) Frontier Research Guarantee INSILICO under Grant EP/Y030494/1, and Engineering and Physical Sciences Research Council (EPSRC) under Grant EP/W007819/1.

\bibliographystyle{model2-names.bst}\biboptions{authoryear}
\bibliography{reference}

\onecolumn

\setcounter{table}{0} 
\setcounter{figure}{0} 
\renewcommand{\thefigure}{\normalsize S\arabic{figure}}
\renewcommand{\thetable}{\normalsize S\arabic{table}}

\makeatletter
\long\def\@makecaption#1#2{%
  \vskip\abovecaptionskip
  \normalsize
  \sbox\@tempboxa{#1: #2}%
  \ifdim \wd\@tempboxa >\hsize
    #1: #2\par
  \else
    \global \@minipagefalse
    \hb@xt@\hsize{\hfil \box\@tempboxa \hfil}%
  \fi
  \vskip\belowcaptionskip
}
\makeatother

\begin{center}

\vspace{12mm}
\Huge{Supplementary Materials}
\vspace{3mm}

\begin{table}[!h]
  \centering
  % \caption{Public BUS dataset introduction.}
  \caption{\normalsize Public BUS dataset introduction.}
  \resizebox{\textwidth}{!}{
    \begin{tabular}{cccccc}
    \toprule
          &  Image\&Mask & Mean Image Size & Mean Nodule Size & Box Area Range & Nodule Area Range \\
    \midrule
    BrEaST & 252   & 523*620 & 136*196 & 2016-217167 & 1515-144084 \\
    STU   & 42    & 402*402 & 149*182 & 546-89794 & 338-58199 \\
    BUSI-Benign & 437   & 495*613 & 106*190 & 828-209121 & 804-209121 \\
    BUSI-malignant & 210   & 494*598 & 206*286 & 1075-244979 & 569-167411 \\
    \bottomrule
    \end{tabular}}%
  \label{tab:intro_pub}%
  \vspace{1em}
\end{table}%

\begin{table}[!h]
  \centering
  \caption{\normalsize Model evaluation on public BUS datasets.}
  \resizebox{\textwidth}{!}{
    \begin{tabular}{ccccccccc}
    \toprule
          & \multicolumn{4}{c}{BrEaST}    & \multicolumn{4}{c}{STU} \\
    \midrule
          & DICE$\uparrow$  & JAC$\uparrow$   & HD$\downarrow$    & ASD$\downarrow$   & DICE$\uparrow$  & JAC$\uparrow$   & HD$\downarrow$    & ASD$\downarrow$ \\
    \midrule
    SAM-Med & 89.16±6.38 & 81.75±9.84 & 25.82±18.17 & 6.34±5.11 & 90.18±6.24 & 82.94±7.12 & 23.46±17.24 & 6.45±3.28 \\
    nnU-Net & 90.26±6.53 & \textbf{84.14±8.02} & 26.85±20.14 & 7.01±6.50 & \textbf{91.65±4.78} & \textbf{84.29±6.01} & 25.88±16.71 & 6.71±4.19 \\
    Ours  & \textbf{90.97±5.12} & 83.83±8.25 & \textbf{19.09±17.64} & \textbf{5.25±4.30} & 91.20±3.56 & 84.01±5.78 & \textbf{18.67±14.12} & \textbf{5.16±2.89} \\
    \midrule
          & \multicolumn{4}{c}{BUSI-benign} & \multicolumn{4}{c}{BUSI-malignant} \\
    \midrule
          & DICE$\uparrow$  & JAC$\uparrow$   & HD$\downarrow$    & ASD$\downarrow$   & DICE$\uparrow$  & JAC$\uparrow$   & HD$\downarrow$    & ASD$\downarrow$ \\
    \midrule
    SAM-Med & 91.63±5.11 & 86.54±8.60 & 19.86±15.49 & 5.41±5.13 & 87.05±7.42 & 78.62±10.18 & 38.63±26.28 & 10.46±8.97 \\
    nnU-Net & 91.21±5.08 & 86.25±8.29 & 20.68±16.11 & 5.30±4.92 & 88.14±5.69 & 80.01±8.53 & 36.74±25.16 & 10.08±7.21 \\
    Ours  & \textbf{91.94±4.68} & \textbf{86.77±7.72} & \textbf{13.84±14.37} & \textbf{3.99±3.86} & \textbf{89.48±4.66} & \textbf{81.29±7.48} & \textbf{29.13±21.05} & \textbf{8.80±5.17} \\
    \bottomrule
    \end{tabular}}%
    \vspace{1em}
  \label{tab:pub_bus}%
\end{table}%

\begin{table}[!h]
  \centering
  \caption{\normalsize Reward analysis on BUS Testing-A and Testing-B datasets. $\Delta$DICE shows the average DICE performance drop.}
  \resizebox{\textwidth}{!}{
    \begin{tabular}{cccccccccccc}
    \toprule
    \multicolumn{3}{c}{\multirow{2}[4]{*}{Reward}} & \multicolumn{4}{c}{BUS (Testing-A)} & \multicolumn{4}{c}{BUS (Testing-B)} & \multirow{3}[6]{*}{$\Delta$DICE$\downarrow$} \\
\cmidrule{4-11}    \multicolumn{3}{c}{}  & \multirow{2}[4]{*}{DICE$\uparrow$} & \multirow{2}[4]{*}{JAC$\uparrow$} & \multirow{2}[4]{*}{HD$\downarrow$} & \multirow{2}[4]{*}{ASD$\downarrow$} & \multirow{2}[4]{*}{DICE$\uparrow$} & \multirow{2}[4]{*}{JAC$\uparrow$} & \multirow{2}[4]{*}{HD$\downarrow$} & \multirow{2}[4]{*}{ASD$\downarrow$} &  \\
\cmidrule{1-3}    CSR   & IDR1  & IDR2  &       &       &       &       &       &       &       &       &  \\
    \midrule
    \ding{51}     & \ding{55}     & \ding{55}     & 91.15±7.17 & 85.99±9.47 & 16.07±18.82 & 6.55±5.79 & 88.14±8.92 & 82.68±10.83 & 17.84±19.05 & 7.23±6.11 & 3.01 \\
    \ding{55}     & \ding{51}     & \ding{55}     & 70.60±13.24 & 61.89±12.92 & 21.78±20.95 & 8.63±7.28 & 68.84±13.76 & 59.76±13.52 & 23.02±21.77 & 9.05±7.95 & 1.76 \\
    \ding{55}     & \ding{55}     & \ding{51}     & 69.92±14.86 & 60.33±13.19 & 22.02±21.38 & 8.51±7.16 & 68.93±14.99 & 59.54±14.12 & 23.18±22.05 & 8.99±8.33 & 0.99 \\
    \ding{55}     & \ding{51}     & \ding{51}     & 71.32±13.72 & 62.44±12.56 & 20.11±19.54 & 8.14±6.94 & 70.53±14.16 & 60.99±13.01 & 21.14±20.20 & 8.53±7.08 & 0.79 \\
    \ding{51}     & \ding{51}     & \ding{55}     & 91.62±6.89 & 85.74±9.18 & 15.87±18.32 & 4.78±5.26 & 91.07±7.85 & 84.83±11.14 & 17.54±19.11 & 5.09±5.35 & 0.55 \\
    \ding{51}     & \ding{55}     & \ding{51}     & 91.71±6.14 & 85.95±7.88 & 15.01±16.87 & 5.09±4.78 & 91.34±5.97 & 85.19±9.74 & 16.61±17.85 & 5.23±5.10 & 0.37 \\
    \ding{51}     & \ding{51}     & \ding{51}     & 92.37±4.22 & 86.09±6.80 & 14.19±13.89 & 3.69±2.50 & 92.15±5.61 & 85.66±9.36 & 16.30±17.40 & 4.07±4.24 & 0.22 \\
    \bottomrule
    \end{tabular}}%
    \vspace{1em}
  \label{tab:reward-AB}%
\end{table}%

 \begin{table}[h]
  \centering
  \small
  \caption{\normalsize Performance of 2D and 3D classifiers using different backbones.}
  \label{tab:cls2}
    \begin{tabular}{cccccccc}
    \toprule
          &       & Accuracy$\uparrow$ & Precision$\uparrow$  & Recall$\uparrow$ & F1-score$\uparrow$ & p-values (Acc) & p-values (F1)\\
    \midrule
    \multirow{3}[2]{*}{2D} & ResNet-18 & 99.167\% & 99.397\% & 98.933\% & 99.165\% & - & - \\
          & ResNet-34 & 99.183\% & 99.697\% & 98.667\% & 99.179\% & $>$0.05 & $>$0.05\\
          & ResNet-50 & 99.200\% & 99.597\% & 98.800\% & 99.190\% & $>$0.05 & $>$0.05 \\
    \midrule
    \multirow{3}[2]{*}{3D} & ResNet-18 & 99.100\% & 98.846\% & 99.360\% & 99.102\% &- &-\\
          & ResNet-34 & 99.120\% & 98.905\% & 99.340\% & 99.122\% & $>$0.05 & $>$0.05\\
          & ResNet-50 & 99.130\% & 99.022\% & 99.240\% & 99.131\% & $>$0.05 & $>$0.05\\
    \bottomrule
    \end{tabular}%
    \vspace{1em}
  \label{tab:addlabel}%
\end{table}%

\begin{table}[h]
  \centering
  \caption{\normalsize Average annotation (Anno.) and testing time for both BUS and ABUS images.}
    \begin{tabular}{cccc}
    \toprule
          & Mask Anno. & Box Anno. & Model Testing \\
    \midrule
    2D (BUS) & 2min  & 10s    & 2s \\
    3D (ABUS) & 25min & 30s   & 15s \\
    \bottomrule
    \end{tabular}%
    \vspace{1em}
\label{tab:time}%
\end{table}%

\begin{figure*}[!h]
	\centering
\includegraphics[width=0.8\linewidth]{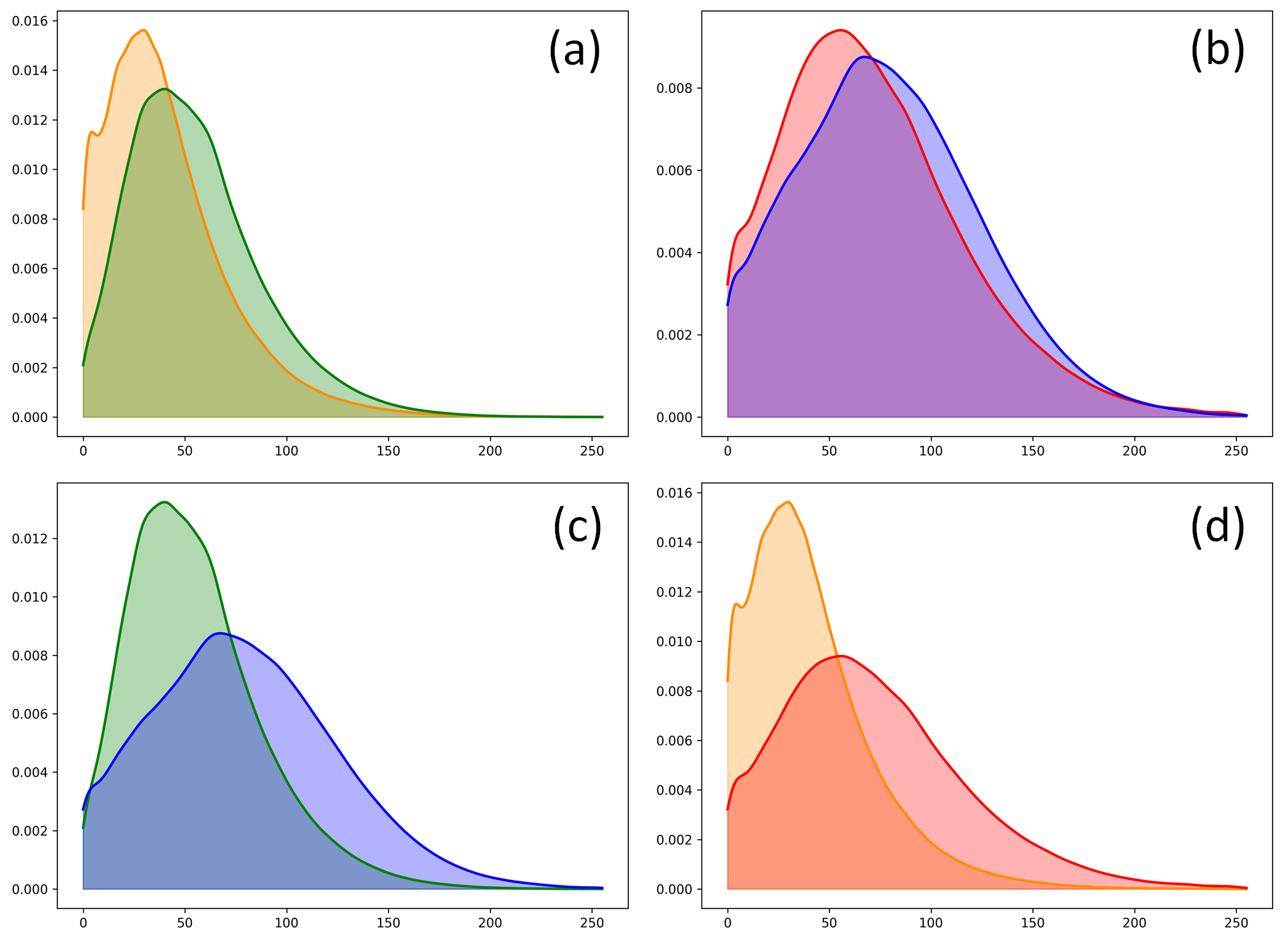}
	\caption{\normalsize Distribution analysis on BUS Testing-A (\textbf{TA}) and Testing-B, (\textbf{TB}). 
    Green\&Yellow curves: foreground distributions on \textbf{TA}\&\textbf{TB}, respectively. 
    Blue\&Red curves: background distributions on \textbf{TA}\&\textbf{TB}, respectively. 
    Sub-figures: foreground (a) and background (b) distributions between two datasets; fore- and back-ground differences within \textbf{TA} (c) and \textbf{TB} (d), respectively.}
    \label{fig:dist_AB}
    % \vspace{3em}
\end{figure*}

\end{center}

\end{document}